\documentclass[journal]{IEEEtai}

\usepackage[colorlinks,urlcolor=blue,linkcolor=blue,citecolor=blue]{hyperref}

\usepackage{color,array}

\usepackage{bbm}
\usepackage[table]{xcolor}
\definecolor{cyan}{rgb}{0.0,0.6,0.9}
\definecolor{darkred}{rgb}{0.6,0.0,0.0}
\definecolor{darkgreen}{rgb}{0.0,0.5,0.0}
\definecolor{darkblue}{rgb}{0.0,0.0,0.5}
\definecolor{notered}{HTML}{d62728}
\definecolor{noteblue}{HTML}{1f77b4}
\definecolor{notegreen}{HTML}{2ca02c}
\definecolor{noteorange}{HTML}{ff7f0e}
\usepackage{soul}

\usepackage{mdframed}
\usepackage{tcolorbox}
\usepackage{multirow} 
\usepackage{booktabs} 
\usepackage{threeparttable}
\usepackage{longtable}
\let\oldnl\nl
\newcommand{\nonl}{\renewcommand{\nl}{\let\nl\oldnl}}
\usepackage{titletoc}
\usepackage{graphicx}
\usepackage{wrapfig}
\usepackage{subcaption}

\usepackage{algorithm}
\usepackage{algorithmicx}
\usepackage{algpseudocode}

\usepackage{caption}

\usepackage{amsmath,amsfonts,bm, amssymb}

\usepackage{amsthm}
\usepackage{mathtools}

\def\1{\bm{1}}

\def\rve{{\mathbf{e}}}

\def\rvg{{\mathbf{g}}}

\def\rvq{{\mathbf{q}}}

\def\rvx{{\mathbf{x}}}

\DeclareMathAlphabet{\mathsfit}{\encodingdefault}{\sfdefault}{m}{sl}
\SetMathAlphabet{\mathsfit}{bold}{\encodingdefault}{\sfdefault}{bx}{n}

\def\gY{{\mathcal{Y}}}

\newcommand{\E}{\mathbb{E}}

\newcommand{\norm}[1]{\left\lVert#1\right\rVert} 
\newcommand{\Norm}[1]{\lVert#1\rVert} 

\newtheorem{assumption}{Assumption}
\newtheorem{definition}{Definition}
\newtheorem{theorem}{Theorem}

\newtheorem{lemma}{Lemma}

\usepackage{orcidlink}

\newcommand{\fl}{\texttt{FL}}

\newcommand{\afl}{\texttt{AFL}}

\begin{document}
	

\title{Asynchronous Federated Learning with non-convex client objective functions and heterogeneous dataset}
\author{Ali Forootani, \IEEEmembership{Senior Member, IEEE}, Raffaele Iervolino, \IEEEmembership{Senior Member, IEEE.}
\thanks{Ali Forootani is with Helmholtz Center for Environmental Research-UFZ, Permoserstrasse 15, 04318 Leipzig, Germany (\texttt{aliforootani@ieee.org/ali.forootani@ufz.de}).}
\thanks{Raffaele Iervolino is with Department of Electrical Engineering and Information Technology, University of Naples, 80125 Napoli, Italy 
 (\texttt{rafierv@unina.it}).}
}


\maketitle

\begin{abstract}
Federated Learning is a distributed machine learning paradigm that enables model training across decentralized devices holding local data, thereby preserving data privacy and reducing the need for centralization. Despite its advantages, traditional FL faces challenges such as communication overhead, system heterogeneity, and straggler effects. Asynchronous Federated Learning has emerged as a promising solution, allowing clients to send updates independently, which mitigates synchronization issues and enhances scalability. This paper extends the Asynchronous Federated Learning framework to scenarios involving clients with non-convex objective functions and heterogeneous dataset, which are prevalent in modern machine learning models like deep neural networks. We provide a rigorous convergence analysis for this setting, deriving bounds on the expected gradient norm and examining the impacts of staleness, variance, and heterogeneity. To address the challenges posed by asynchronous updates, we introduce a staleness-aware aggregation mechanism that penalizes outdated updates, ensuring fresher data has a more significant influence on the global model. Additionally, we propose a dynamic learning rate schedule that adapts to client staleness and heterogeneity, improving stability and convergence.

Our approach effectively manages heterogeneous environments, accommodating differences in client computational capabilities, data distributions, and communication delays, making it suitable for real-world Federated Learning applications. We also analyze the effects of client selection methods—specifically, choosing clients with or without replacement—on variance and model convergence, providing insights for more effective sampling strategies. The practical implementation of our methods using \texttt{PyTorch} and \texttt{Python}'s asyncio library demonstrates their applicability in real-world asynchronous and heterogeneous FL scenarios. Empirical experiments validate the proposed methods, showing improved performance and scalability in handling asynchronous updates, and non-convex client's objective function with associated heterogeneous dataset.
\end{abstract}

\begin{IEEEImpStatement}
	Asynchronous Federated Learning addresses key challenges in distributed learning, yet its application to non-convex optimization remains underexplored. This work extends Asynchronous Federated Learning to non-convex client objectives and heterogeneous datasets, providing a rigorous convergence analysis and quantifying the effects of staleness and variance. We introduce a staleness-aware aggregation mechanism and a dynamic learning rate schedule, improving stability and convergence. Empirical results on MNIST and CIFAR-10 demonstrate superior scalability and robustness over synchronous Federated Learning. Implemented using \texttt{PyTorch} and ``asyncio'', our approach enhances real-world applicability in privacy-sensitive domains like healthcare and finance. This work significantly advances Asynchronous Federated Learning's theoretical foundations and practical deployment, enabling more efficient, decentralized learning.
\end{IEEEImpStatement}

\begin{IEEEkeywords}
Federated Learning, Stochastic Gradient Descent, Client Drifts, Asynchronous Federated Learning.
\end{IEEEkeywords}

\section{Introduction}
\IEEEPARstart{F}{e}derated Learning (FL) is a distributed machine learning paradigm that enables the training of models across multiple decentralized devices or servers holding local data samples, without requiring the exchange of the data itself \cite{yu2019parallel, kairouz2021advances}. This approach addresses privacy concerns and reduces the need for extensive data centralization \cite{mcmahan2017communication}. FL has become increasingly relevant with the growing ubiquity of edge devices and the rising importance of data privacy regulations \cite{bonawitz2019towards, mothukuri2021survey}. Its applications span various domains, including healthcare \cite{xu2021federated}, finance \cite{long2020federated}, renewable energies \cite{forootani2024asynchronous}, and mobile computing \cite{yu2021toward}.

Despite its potential, traditional FL faces significant challenges. These include communication overhead, system heterogeneity, and straggler effects \cite{li2020federated, wang2019adaptive}. Communication overhead arises due to frequent exchanges of model parameters between clients and the central server. System heterogeneity reflects the diversity in computation power, network connectivity, and local data distribution across devices \cite{sattler2020sparse, pang2020realizing}. The straggler effect occurs when slower clients delay the global model aggregation process, impeding efficiency \cite{harlap2016addressing, cipar2013solving, chai2020fedat}.

Asynchronous Federated Learning (\afl) has emerged as a promising solution to mitigate these challenges \cite{chen2020asynchronous}. Unlike synchronous FL, where all participating devices must complete their local training before model aggregation, \afl~allows clients to send updates to the central server independently and at different times \cite{hsieh2020non}. This flexibility reduces waiting times, improves scalability, and accommodates devices with varying computational capabilities. Furthermore, \afl~inherently addresses the straggler problem by not requiring synchronization across all clients, leading to faster convergence and more efficient resource utilization \cite{xie2020asynchronous}.

\subsection{Literature Review}

Federated Learning has garnered significant attention since its inception. Federated Averaging (FedAvg) algorithm is introduced in \cite{mcmahan2017communication}, which became a cornerstone in FL research. They demonstrated the feasibility of FL in training deep learning models while preserving user privacy. However, subsequent studies highlighted its limitations. For example in \cite{bonawitz2019towards} a secure aggregation techniques to enhance privacy in FL has been explored, while the work reported in \cite{karimireddy2020scaffold} proposed adaptive optimizers to tackle the challenges posed by non-IID (non-Independent and Identically Distributed) data distributions across clients.

Addressing communication bottlenecks has been another focus of research. For example in \cite{sattler2020sparse} the sparse updates and compression techniques has been introduced to reduce the volume of data exchanged between clients and the server.

In another line of research the use of personalization layers in FL to accommodate diverse client data distributions without excessive communication was discussed \cite{li2020federated}.

The concept of asynchronous FL was first explored in \cite{hsieh2020non}, who proposed FedAsync, an \afl~approach that allows updates from clients to be aggregated as they arrive. This method showed promise in mitigating the straggler problem and improving convergence speed. Subsequent works, such as \cite{xie2020asynchronous}, investigated the robustness of \afl~under varying client participation rates and proposed strategies to balance stale updates.


While \afl~addresses many challenges, it introduces new complexities, such as handling stale updates and ensuring fairness in model aggregation. Several works have advanced the field of asynchronous decentralized and parallel stochastic gradient descent (SGD), addressing challenges such as gradient staleness, convergence, and scalability. For example AD-PSGD was introduced in \cite{lian2017asynchronous, wang2021adaptive}, which removes the central server bottleneck while maintaining optimal convergence rates despite staleness. Adaptive frameworks like MindTheStep-AsyncPSGD adjusts step sizes to mitigate the impact of stale gradients \cite{backstrom2022mindthestep}, while instance-based Adaptiveness to Staleness in Asynchronous SGD has been considered to improve performance \cite{backstrom2022asap}.

Theoretical analyses, such as those reported in \cite{dutta2018slow}, highlight error-runtime trade-offs and scenarios where stale gradients can be advantageous \cite{dutta2018slow}. Techniques like delay compensation effectively address delays in asynchronous SGD \cite{zheng2017delay}, while the parallelization limits under gradient staleness has been explored in \cite{zhang2022parallelization}. Acceleration methods have also been integrated into asynchronous SGD to enhance convergence rates \cite{async2022accelerated}. Studies reported in \cite{zhou2018unbounded}, investigated the tolerable limits of unbounded delays in distributed optimization, while in \cite{lian2015nonconvex}, the properties of asynchronous SGD in non-convex settings has been investigated. In \cite{keuper2015asynchronous}, an asynchronous communication algorithm to address stale gradients in parallel updates was proposed \cite{keuper2015asynchronous}.

To address the challenges posed by stragglers in synchronous federated learning (SFL) and improve its efficiency, several \afl~schemes have been proposed \cite{chen2020asynchronous,gu2021privacy,AsyncFed,chen2018lag}. \cite{wu2020safa} introduced the Semi-Asynchronous Federated Learning Algorithm (SAFA), which classifies clients into three categories—sustainable, moderate, and unsustainable—allowing only sustainable clients to operate asynchronously, thereby reducing delays and enhancing system efficiency.
In \cite{FedBuff} a buffering mechanism has been proposed where the server aggregates local updates once a predefined buffer size is reached, balancing asynchronous updates with consistency in global model performance. Meanwhile, \cite{FedSA} developed a two-stage training process that assigns each client a staleness parameter during the convergence phase, ensuring updates arrive synchronously despite varying processing speeds, thereby maintaining robust convergence in asynchronous settings. These strategies collectively mitigate the straggler problem while preserving the effectiveness and scalability of federated learning.


Our previous work \cite{forootani2024asynchronous}, focused on \afl~with convex assumption on local cost functions. In this work, we extend this approach to clients with non-convex objective functions. This extension is significant since non-convex optimization problems are prevalent in machine learning, especially in training deep neural networks. Recent literature has also explored \afl~with non-convex objectives. For instance, in \cite{xie2019asynchronous} an asynchronous federated optimization algorithm was proposed and its convergence for both strongly convex and a restricted family of non-convex problems have been proved. Another relevant work has studied improved convergence rates for asynchronous stochastic gradient descent in non-convex settings \cite{koloskova2022sharper}.

This body of literature underscores the evolution of federated learning from its synchronous roots to asynchronous implementations. \afl~represents a crucial step forward in addressing the inherent limitations of traditional FL, particularly in scenarios involving diverse and large-scale distributed systems. However, further research is needed to refine its algorithms, enhance robustness, and validate its efficacy across diverse real-world applications.

\subsection{Contributions}

\begin{itemize}
    \item This paper extends the theoretical and practical framework of asynchronous federated learning to \textit{non-convex objective functions}, addressing the challenges of multiple local minima and saddle points common in modern machine learning models.

    \item We provide a rigorous \textit{convergence analysis} for non-convex asynchronous federated learning, deriving bounds on the expected gradient norm and analyzing the impact of key factors such as \textit{staleness}, \textit{variance}, and \textit{heterogeneity}.

    \item We introduce and formalize the concept of \textit{drift} due to staleness in asynchronous updates, quantifying its impact on the global model and providing mathematical bounds to mitigate its effects.

    \item We propose \textit{staleness-aware aggregation mechanism}, penalizing updates based on their staleness to ensure that fresher updates contribute more significantly to the global model.

    \item We introduce a \textit{dynamic learning rate schedule} that adapts to client staleness and heterogeneity, improving stability and convergence in asynchronous federated learning.

    \item Our approach effectively handles \textit{heterogeneous environments}, addressing differences in client computation capabilities, data distributions, and communication delays, making it suitable for real-world federated learning applications.

    \item We analyze the effect of \textit{choosing clients with replacement} vs. \textit{choosing clients without replacement} on variance and model convergence, providing insights for more effective sampling in federated learning.
    
\end{itemize}

We emphasize a practical implementation using \texttt{PyTorch} and Python's \texttt{asyncio} library, demonstrating real-world applicability for asynchronous and heterogeneous federated learning scenarios. We validate the proposed methods through \textit{empirical experiments}, demonstrating improved performance and scalability in handling asynchronous updates and non-convex objectives.

This paper is organized as follows: in Section \ref{preliminaries} preliminaries are discussed. Problem formulation and an overview of the proposed \afl~Algorithm is given in Section \ref{problem_formulation}. In section \ref{converegence_analysis} we will provide the convergence analysis of the \afl~Algorithm with non-convex client objective function and heterogeneous dataset. We evaluate the \afl~Algorithm and compare its results with synchronous FL on well known dataset such as MNIST and CIFAR-10 in Section \ref{simulation}. The paper ends with Conclusion in Section \ref{conclusion}.


\section{Preliminaries}\label{preliminaries}
In this section, we establish the necessary notation, definitions, and assumptions required for analyzing the \afl~framework. We begin by introducing the mathematical notation used throughout this paper, followed by a discussion of different sampling methods and their statistical properties. Next, we present key theoretical results related to variance bounds and martingale properties, which form the foundation for our analysis. Finally, we introduce key assumptions that characterize the behavior of local objective functions, stochastic gradients, and model drift in the presence of stale updates.

\subsection{Notation}
Let $\norm{\cdot}$ denote the standard Euclidean norm, applicable to both vectors and matrices. The parameter $\sigma^2$ provides an upper bound on the variance of the stochastic gradients for individual clients participating in the \afl~process. The total number of clients is represented by $C$, with the index $c$ referring to a specific client, and $J$ indicating the subset of clients actively engaged in training during a particular round. Training proceeds over $\mathcal{J}$ global communication rounds, indexed by $j$, and each client performs $I$ local update steps per round, indexed by $i$. The learning rate or step size is denoted by $\lambda$, and the effective learning rate in the \afl~framework is defined as $\tilde\lambda = \lambda C I$.

A permutation of the client indices is represented by $\psi$, defined as $\{\psi_1, \psi_2, \ldots, \psi_C\}$, which maps the set $\{1, 2, \ldots, C\}$ to a reordered sequence. The global objective function is denoted by $\mathcal{L}$, while the local objective function for client $c$ is expressed as $\mathcal{L}_c$. The global model parameters at the $j$-th communication round are denoted by $\theta^{(j)}$, and the local model parameters of client $c$ after $i$ local update steps within the $j$-th round are represented by $\theta_{c,i}^{(j)}$. The stochastic gradients corresponding to the local objective function $\mathcal{L}_{\psi_c}$ of client $\psi_c$, evaluated at $\theta_{c,i}^{(j)}$, are denoted by $\rvq_{\psi_c,i}^{(j)}$, where $\rvq_{\psi_c,i}^{(j)} \coloneqq \nabla l_{\psi_c}(\theta_{c,i}^{(j)}; \xi)$, with $\xi$ representing a stochastic sampling variable.

\subsection{Reviews on some probability and stochastic concepts}
We analyze the properties of the sample mean \( \overline{\theta}_\psi \) under two common sampling approaches: (i) sampling with replacement and (ii) sampling without replacement. First, the population mean \( \overline{\theta} \) and variance \( \nu^2 \) are defined as:
\[
\overline{\theta} \coloneqq \frac{1}{m} \sum_{k=1}^m \theta_k, \quad \nu^2 \coloneqq \frac{1}{m} \sum_{k=1}^m \norm{\theta_k - \overline{\theta}}^2,
\]
\noindent where \( \theta_1, \theta_2, \dots, \theta_m \) represent the fixed vectors in the population. Given a sample of \( s \) vectors, denoted as \( \rvx_{\psi_1}, \rvx_{\psi_2}, \dots, \rvx_{\psi_s} \), drawn from this population, the sample mean is expressed as:
\[
\overline{\theta}_\psi = \frac{1}{s} \sum_{p=1}^s \theta_{\psi_p}.
\]
Below, we evaluate the expected value and variance of \( \overline{\theta}_\psi \) under the two sampling methods:
\paragraph{Sampling with Replacement}
(i) Expected Value : The expected value of the sample mean equals the population mean \cite{forootani_afl_emt}:
  \[
  \E[\overline{\rvx}_\psi] = \overline{\theta}.
  \]
(ii) Variance: The variance of the sample mean is given by:
  \[
  \E\left[\norm{\overline{\rvx}_\psi - \overline{\theta}}^2\right] = \frac{\nu^2}{s}.
  \]
  This result shows that the variance decreases inversely with the sample size \( s \) but is independent of the total population size \( m \).

\paragraph{Sampling without Replacement}
(i) Expected Value: The expected value of the sample mean remains equal to the population mean \cite{forootani_afl_emt}: \( \E[\overline{\theta}_\psi] = \overline{\theta}.\)
(ii) Variance: The variance in this case is: \(
  \E\left[\norm{\overline{\rvx}_\psi - \overline{\theta}}^2\right] = \frac{m - s}{s (m - 1)} \nu^2.
  \)
  Here, the variance depends on both the sample size \( s \) and the total population size \( m \), and it is typically smaller than the variance in the case of sampling with replacement.

In summary, while the expected value of the sample mean is identical for both sampling methods, the variance differs, with sampling without replacement usually leading to a smaller variance.

Consider a sequence of random variables \(\{\epsilon_i\}_{i=1}^m\) and associated random vectors \(\{\theta_i\}_{i=1}^m\), where each \(\theta_i \in \mathbb{R}^d\) depends on the history \(\epsilon_1, \epsilon_2, \ldots, \epsilon_i\). Assume the conditional expectation satisfies:
\[
\mathbb{E}_{\epsilon_i}[\theta_i \mid \epsilon_1, \ldots, \epsilon_{i-1}] = \rve_i,
\]
\noindent which implies that \(\{\theta_i - \rve_i\}_{i=1}^m\) forms a martingale difference sequence with respect to the filtration generated by \(\{\epsilon_i\}_{i=1}^m\). Additionally, assume that the conditional variance of each \(\theta_i - \rve_i\) is uniformly bounded:
\[
\mathbb{E}_{\epsilon_i}[\|\theta_i - \rve_i\|^2 \mid \epsilon_1, \ldots, \epsilon_{i-1}] \leq \delta^2,
\]
\noindent for some constant \(\delta > 0\). Then, the total variance of the sum of the martingale difference sequence satisfies:
\begin{equation}\label{martingle}
\mathbb{E}\left[\left\|\sum_{k=1}^m (\theta_k - \rve_k)\right\|^2\right] = \sum_{k=1}^m \mathbb{E}\left[\|\theta_k - \rve_k\|^2\right] \leq m\delta^2.  
\end{equation}

This inequality demonstrates that the variance of the sum is bounded by \( m\delta^2 \), where \( m \) is the number of terms in the sequence and \( \delta^2 \) is the uniform variance bound for each term.

Using the approach of sampling clients without replacement, define \cite{karimireddy2020scaffold, forootani_afl_emt}:

\[
p_{c,i}(k) = 
\begin{cases}
I-1, & \text{if } k \leq c-1, \\
i-1, & \text{if } k = c,
\end{cases}
\]

\noindent where \( I \) is a fixed integer. For \( J \leq C \) and \( C \geq 2 \), the following inequality holds \cite{karimireddy2020scaffold, forootani_afl_emt}:

\[
\sum_{c=1}^J \sum_{i=0}^{I-1} \mathbb{E} \left\|\sum_{k=1}^c \sum_{j=0}^{p_{c,i}(k)} (\theta_{\psi_k} - \overline{\theta}) \right\|^2 \leq \frac{1}{2} J^2 I^3 \nu^2,
\]

\noindent where \( \overline{\theta} \) is the population mean, and \( \nu^2 \) is the population variance.


\subsection{Definitions and main assumptions}

In this subsection we provide definitions, and main assumptions that are required for the convergence proof of the \afl~algorithm.

\begin{definition}  
In \afl~, each client updates its local model based on a potentially stale global model. This means that client updates might not align with the current global model, leading to a drift between the models at each client and the global model. Let’s define this drift mathematically:

   - At time \( t \), the global model is \( \theta^{(t)} \), and client \( i \) uses a stale model \( \theta^{(t-\tau_i)} \) (with \( \tau_i \) being the staleness of client \( i \)'s model).
   - The drift for client \( i \) at time \( t \) can be expressed as:
     \[
     \Delta \theta_i^{\text{drift}} = \theta^{(t)} - \theta^{(t-\tau_i)}.
     \]

\end{definition}

\paragraph{Impact of Drift on Updates} This drift can affect the quality of local updates. The local gradient computed by client \( i \) is based on a stale model \( \theta^{(t-\tau_i)} \), and the direction of the gradient will be influenced by the mismatch between \( \theta^{(t)} \) and \( \theta^{(t-\tau_i)} \).

\begin{assumption}\label{l_smooth}
Each local objective \( \mathcal{L}_i(\theta) \) is \( L \)-smooth:
\[
\mathcal{L}_i(\theta_2) \leq \mathcal{L}_i(\theta_1) + \langle \nabla \mathcal{L}_i(\theta_1), \theta_2 - \theta_1 \rangle + \frac{L}{2} \|\theta_2 - \theta_1\|^2.
\]
\end{assumption}

\begin{assumption}\label{bounded_variance}
The variance of the stochastic gradients is bounded:
\[
\mathbb{E}\left[\|\nabla \mathcal{L}_i(\theta) - \nabla \mathcal{L}(\theta)\|^2\right] \leq \sigma^2.
\] 
\end{assumption}

\begin{assumption}\label{max_delay}
The staleness \( \tau_i \) is bounded by \( \tau_{\max} \):
\[
\tau_i \leq \tau_{\max}.
\] 
\end{assumption}

\begin{assumption}\label{bounded_drift}
The drift between the local and global models is bounded:
\[
\|\Delta \theta_i^{\text{drift}}\| \leq \Delta \theta_{\text{max}},
\]
where \( \Delta \theta_{\text{max}} \) represents the maximum allowable drift between the global model and the local model due to staleness.
 
\end{assumption}

\begin{assumption}
Local updates are unbiased estimates of the true gradient:
\[
\mathbb{E}[\nabla \mathcal{L}_i(\theta)] = \nabla \mathcal{L}_i(\theta).
\]
\end{assumption}

\begin{assumption}
Clients are randomly selected at each round, and each client computes the update using the local model \( \theta^{(t-\tau_i)} \).
\end{assumption}

\begin{assumption}\label{asm:heterogeneity:optima}
We assume that, at the global minimizer \( \theta^\ast \), the gradients of the individual local loss functions are closely aligned. Specifically, there exists a constant \( \beta_\ast^2 \) such that:  
\[
\frac{1}{C} \sum_{c=1}^C \norm{\nabla \mathcal{L}_c(\theta^\ast)}^2 = \beta_\ast^2,
\]
where \( \beta_\ast \) measures the degree of similarity among the gradients at \( \theta^\ast \), the solution to the global loss \( \mathcal{L}(\theta) \).
\end{assumption}


\section{Problem Formulation}\label{problem_formulation}

The objective of the \fl~framework is to optimize a global loss function, defined as:  
\[
\min_{\theta \in \mathbb{R}^d} \left\{ \mathcal{L}(\theta) \coloneqq \frac{1}{C} \sum_{c=1}^{C} \mathcal{L}_c(\theta) \right\},
\]
where \( \mathcal{L}_c(\theta) \coloneqq \mathbb{E}_{\xi \sim \mathcal{D}_c} [\ell_c(\theta; \xi)] \) represents the local objective function for client \( c \). Here, \( \ell_c(\theta; \xi) \) denotes the client-specific loss function evaluated at model parameters \( \theta \) for data point \( \xi \), \( \mathcal{D}_c \) is the data distribution of client \( c \), and \( C \) is the total number of clients.

If \( \mathcal{D}_c \) consists of a finite dataset, \( \mathcal{D}_c = \{\xi_c^j \mid j = 1, 2, \dots, |\mathcal{D}_c|\} \), the local objective function can be expressed equivalently as:  
\[
\mathcal{L}_c(\theta) = \frac{1}{|\mathcal{D}_c|} \sum_{j=1}^{|\mathcal{D}_c|} \ell_c(\theta; \xi_c^j).
\]

In federated learning, clients operate independently and often experience asynchronous delays in their updates. The delay for client \( c \) is denoted as \( \tau_c \), representing the time lag between the computation of the local gradient by the client and the incorporation of the gradient into the global model.

During each training round \( t \), a subset of clients \( J^{(t)} = \{c_1, c_2, \dots, c_k\} \subseteq \{1, 2, \dots, C\} \) is selected at random without replacement to participate in training. For each client \( c \in J^{(t)} \), the process is as follows:

\begin{enumerate}
    \item Model Initialization: The client initializes its local model parameters using the global parameters available at the time, \( \theta^{(t - \tau_c)} \), where \( \tau_c \) accounts for the update delay.

    \item Local Training: The client performs \( I \) iterations of stochastic gradient descent (SGD) on its local dataset, producing intermediate parameter states \( \theta_{c,i}^{(t)} \) for \( i = 0, 1, \dots, I \).

    \item Parameter Transmission: After completing local updates, the client sends its final parameters, \( \theta_{c,I}^{(t)} \), to the central server.
    
\end{enumerate}

The local parameter updates at each iteration \( i \) are computed as:  
\[
\theta_{c,i+1}^{(t)} = \theta_{c,i}^{(t)} - \eta \rvq_{c,i}^{(j)},
\]
where \( \eta > 0 \) is the learning rate, and \( \rvq_{c,i}^{(j)}= \nabla \ell_{c}(\theta_{c,i}^{(t)}; \xi) \) is the stochastic gradient of the local loss with respect to \( \theta_{c,i}^{(t)} \), computed using a randomly sampled data point \( \xi \sim \mathcal{D}_c \). The initialization for local training is given by:  
\[
\theta_{c,0}^{(t)} = \theta^{(t - \tau_c)}.
\]

Once all participating clients \( c \in J^{(t)} \) complete their local training and transmit their updated parameters to the central server, the global model is updated using an aggregation function:  
\[
\theta^{(t+1)} = \text{Aggregate}\left(\{\theta_{c,I}^{(t)} \mid c \in {J}^{(t)}\}\right),
\]
where the aggregation function may involve techniques such as weighted averaging based on the size of local datasets or other strategies to mitigate the effects of heterogeneity across clients.

This iterative process is repeated over multiple rounds, progressively refining the global model parameters \( \theta \) based on asynchronous updates from the distributed clients. The federated learning framework thus enables collaborative training while preserving data privacy and accommodating communication and computation constraints at the client level. Algorithm \ref{afl_non_convex_alg} provides an overview of the proposed \afl.

	\begin{algorithm}
		\caption{Federated Learning with Asynchronous Client Updates}
        \label{afl_non_convex_alg}
		\begin{algorithmic}[1]
			\State \textbf{Input:} Learning rate \( \eta \), number of local iterations \( I \), number of communication rounds \( \mathcal{J} \), convergence threshold \( \kappa \).
			\State \textbf{Initialize:} Global model \( \theta^{(0)} \).
			
			\For{$t = 0, 1, \dots, \mathcal{J}-1$}
			\State Randomly select a subset of clients \( J^{(t)} \subseteq \{1, 2, \dots, C\} \).
			\For{each client \( c \in J^{(t)} \) \textbf{in parallel}}
			\State Initialize local model using the delayed global model:
			\[
			\theta_{c,0}^{(t)} = \theta^{(t - \tau_c)},
			\]
			where \( \tau_c \) is the staleness of client \( c \).
			\For{$i = 0, 1, \dots, I-1$}
			\State Perform local update using stochastic gradient descent:
			\[
			\theta_{c,i+1}^{(t)} = \theta_{c,i}^{(t)} - \eta \nabla \ell_c(\theta_{c,i}^{(t)}; \xi),
			\]
			where \( \xi \sim \mathcal{D}_c \) is a randomly sampled data point.
			\EndFor
			\State Send updated parameters \( \theta_{c,I}^{(t)} \) and staleness \( \tau_c \) to the server.
			\EndFor
			
			\State Aggregate updates to compute the new global model:
			\[
			\theta^{(t+1)} = \text{Aggregate}\left(\{\theta_{c,I}^{(t)} \mid c \in J^{(t)}\}\right).
			\]
			
			\If{$\|\theta^{(t+1)} - \theta^{(t)}\| \leq \kappa$}
			\State \textbf{Break.}
			\EndIf
			\EndFor
			\State \textbf{Output:} Final global model \( \theta^{(\mathcal{J})} \).
		\end{algorithmic}
	\end{algorithm}

\section{Convergence Analysis}\label{converegence_analysis}
In this section, we establish theoretical guarantees on the expected improvement in the loss function across training rounds. Specifically, we derive a recursive bound that characterizes the evolution of the model parameters and provides insights into how step size, variance, and system heterogeneity impact convergence. We first present a key recursion bound in Lemma~\ref{lem:non-convex:recursion}, which quantifies the expected descent in the loss function over a single training round. This result is further complemented by Lemma~\ref{lem:non-convex:drift}, where we analyze the drift in model updates due to the decentralized nature of the optimization process. These results collectively provide a rigorous foundation for understanding the convergence properties of our algorithm in the presence of non-convex objectives.

\begin{lemma}[Non-Convex Recursion Bound]\label{lem:non-convex:recursion}
	Under Assumptions~\ref{l_smooth}, \ref{bounded_variance}, and \ref{asm:heterogeneity:optima}, if the step size \( \gamma \) satisfies
	\[
	\gamma \leq \frac{1}{6LJI(1 + \beta^2/J)},
	\]
	then the following holds for the expected improvement in the loss function \( \mathcal{L} \) after one training round:
	\begin{align*}
		\E\left[\mathcal{L}(\theta^{(j+1)}) - \mathcal{L}(\theta^{(j)})\right] &\leq -\frac{1}{6}JI\gamma \E\left[\|\nabla \mathcal{L}(\theta^{(j)})\|^2\right] \\
		& + 2L\gamma^2JI \sigma^2 \\ 
        & + 2L\gamma^2J^2I^2\frac{C-J}{J(C-1)}\nu^2 \\
		& + \frac{5}{6}L^2\gamma \sum_{c=1}^J \sum_{i=0}^{I-1} \E\left[\|\theta_{c,i}^{(j)} - \theta^{(j)}\|^2\right].
	\end{align*}
\end{lemma}

\begin{proof}

We focus on a single training round and omit the superscript \( j \) for simplicity. All expectations are conditioned on \( \theta^{(j)} \), unless otherwise stated. Since \( \mathcal{L} \) is \( L \)-smooth (Assumption~\ref{l_smooth}), we begin with:
\[
\E\left[\mathcal{L}(\theta + \Delta \theta) - \mathcal{L}(\theta)\right] \leq \E\left[\langle \nabla \mathcal{L}(\theta), \Delta \theta \rangle\right] + \frac{L}{2} \E\|\Delta \theta\|^2.
\]

We first bound \( \E\left[\langle \nabla \mathcal{L}(\theta), \Delta \theta \rangle\right] \). Substitute the overall update \( \Delta \theta \) into the inner product term:
\begin{multline}
\E\left[ \langle \nabla \mathcal{L}(\theta), \Delta \theta \rangle\right] = -\gamma J I \E\Bigg[\Big\langle \nabla \mathcal{L}(\theta)\\, \frac{1}{J} \sum_{c=1}^J \frac{1}{I} \sum_{i=0}^{I-1} \big(\nabla \mathcal{L}_{\pi_c}(\theta_{c,i}) - \nabla \mathcal{L}_{\pi_c}(\theta) + \nabla \mathcal{L}_{\pi_c}(\theta)\big)\Big\rangle\Bigg].
\end{multline}
\noindent Simplifying the expression:
\begin{multline*}
\E\left[\langle \nabla \mathcal{L}(\theta), \Delta \theta \rangle\right] = -\gamma JI \|\nabla \mathcal{L}(\theta)\|^2 \\ - \gamma JI \E\Big[\Big\langle \nabla \mathcal{L}(\theta), \frac{1}{J} \sum_{c=1}^J \frac{1}{I} \sum_{i=0}^{I-1} \Big(\nabla \mathcal{L}_{\pi_c}(\theta_{c,i}) - \nabla \mathcal{L}_{\pi_c}(\theta)\Big)\Big\rangle\Big].  
\end{multline*}

\noindent Using the inequality \( |\langle a, b \rangle| \leq \frac{1}{2} \|a\|^2 + \frac{1}{2} \|b\|^2 \), we obtain:
\begin{multline*}
\E\left[\langle \nabla \mathcal{L}(\theta), \Delta \theta \rangle\right] \leq -\gamma JI \|\nabla \mathcal{L}(\theta)\|^2 + \frac{1}{2} \gamma JI \|\nabla \mathcal{L}(\theta)\|^2 \\ + \frac{1}{2} \gamma \sum_{c=1}^J \sum_{i=0}^{I-1} \E\|\nabla \mathcal{L}_{\pi_c}(\theta_{c,i}) - \nabla \mathcal{L}_{\pi_c}(\theta)\|^2.
\end{multline*}

Using the smoothness assumption (Assumption~ \eqref{l_smooth}):
\begin{multline}
\E\left[\langle \nabla \mathcal{L}(\theta), \Delta \theta \rangle\right] \leq -\frac{1}{2} \gamma JI \|\nabla \mathcal{L}(\theta)\|^2 \\+ \frac{1}{2} L^2 \gamma \sum_{c=1}^J \sum_{i=0}^{I-1} \E\|\theta_{c,i} - \theta\|^2.    
\end{multline}

Next, we bound \( \E\|\Delta \theta\|^2 \) as follows:
\begin{multline}\label{eq:recursion-non-convex-1}
\frac{1}{2} L \E\|\Delta \theta\|^2 \leq 2L\gamma^2 \E\left\|\sum_{c=1}^J \sum_{i=0}^{I-1} \left(\rvg_{\pi_c,i} - \nabla \mathcal{L}_{\pi_c}(\theta_{c,i})\right)\right\|^2 \\+ 2L\gamma^2 \E\left\|\sum_{c=1}^J \sum_{i=0}^{I-1} \left(\nabla \mathcal{L}_{\pi_c}(\theta_{c,i})  - \nabla \mathcal{L}_{\pi_c}(\theta)\right)\right\|^2
\\+ 2L\gamma^2 \E\left\|\sum_{c=1}^J \sum_{i=0}^{I-1} \left(\nabla \mathcal{L}_{\pi_c}(\theta) - \nabla \mathcal{L}(\theta)\right)\right\|^2 \\+ 2L\gamma^2 \E\left\|\sum_{c=1}^J \sum_{i=0}^{I-1} \nabla \mathcal{L}(\theta)\right\|^2.
\end{multline}

\noindent We now bound each term on the right-hand side \eqref{eq:recursion-non-convex-1}, for the first term we have
\[
2L\gamma^2 \E\left\|\sum_{c=1}^J \sum_{i=0}^{I-1} \left(\rvg_{\pi_c,i} - \nabla \mathcal{L}_{\pi_c}(\theta_{c,i})\right)\right\|^2 \leq 2L\gamma^2 JI \sigma^2. \tag{see \eqref{martingle}, Assumption~\ref{bounded_variance}}
\]
\noindent For the second term we write from Assumption \ref{l_smooth}:

\begin{multline*}
2L\gamma^2 \E\left\|\sum_{c=1}^J \sum_{i=0}^{I-1} \left(\nabla \mathcal{L}_{\pi_c}(\theta_{c,i}) - \nabla \mathcal{L}_{\pi_c}(\theta)\right)\right\|^2 \\ \leq 2L^3 \gamma^2 JI \sum_{c=1}^J \sum_{i=0}^{I-1} \E\|\theta_{c,i} - \theta\|^2. 
\end{multline*}

\noindent For the third and forth term we get:
\begin{multline}
2L\gamma^2 \E\left\|\sum_{c=1}^J \sum_{i=0}^{I-1} \left(\nabla \mathcal{L}_{\pi_c}(\theta) - \nabla \mathcal{L}(\theta)\right)\right\|^2 \\+ 2L\gamma^2 \E\left\|\sum_{c=1}^J \sum_{i=0}^{I-1} \nabla \mathcal{L}(\theta)\right\|^2 \leq 2L\gamma^2 J^2I^2 \tfrac{C-J}{J(C-1)}\nu^2 \\ + 2L\gamma^2 J^2I^2 \left(1 + \tfrac{C-J}{J(C-1)}\right) \|\nabla \mathcal{L}(\theta)\|^2.    
\end{multline}

\noindent Substituting the above bounds and applying \( \gamma \leq \frac{1}{6LJI (1 + \beta^2/J)} \), we recover the superscripts and obtain:

\begin{multline}
\E\left[\mathcal{L}(\theta^{(j+1)}) - \mathcal{L}(\theta^{(j)})\right] \leq -\frac{1}{6} JI\gamma \E\|\nabla \mathcal{L}(\theta^{(j)})\|^2 \\+ 2L\gamma^2 JI \sigma^2 + 2L\gamma^2 J^2I^2 \tfrac{C-J}{J(C-1)} \nu^2 \\ + \frac{5}{6} L^2 \gamma \sum_{c=1}^J \sum_{i=0}^{I-1} \E\|\theta_{c,i}^{(j)} - \theta^{(j)}\|^2.   
\end{multline}
	
\end{proof}

\begin{lemma}\label{lem:non-convex:drift}
	Suppose Assumptions~\ref{l_smooth}, \ref{bounded_variance}, \ref{asm:heterogeneity:optima} are satisfied. If $\gamma \leq \frac{1}{6LIJ(1+\beta^2/J)}$, then
	\begin{align*}
		E_j &\leq \frac{9}{4} J^2 I^2 \gamma^2 \nu^2 + \frac{9}{4} J^2 I^3 \gamma^2 \mu^2 \\ &+ \frac{9}{4} \left( \frac{\beta^2}{J} + 1 \right) J^3 I^3 \gamma^2 \mathbb{E} \left[ \left\| \nabla \mathcal{L}(\theta^{(j)}) \right\|^2 \right].
	\end{align*}
\end{lemma}

\begin{proof}
	According to Algorithm~\ref{afl_non_convex_alg}, the overall updates of \afl~from $\theta^{(j)}$ to $\theta_{c,i}^{(j)}$ are
	\begin{align*}
		\theta_{c,i}^{(j)} - \theta^{(j)} = -\gamma \sum_{k=1}^{c} \sum_{f=0}^{p_{c,i}(k)} \rvq_{\psi_k,f}^{(j)},\\ \quad p_{c,i}(k) = 
		\begin{cases}
			I-1, & \text{if } k \leq c-1, \\
			i-1, & \text{if } k = c.
		\end{cases}
	\end{align*}
	
	\noindent For the sake of simplicity we consider only one training round and we drop our dependency to superscripts $j$. Unless otherwise stated, the expectation is conditioned on $\theta^{(j)}$.
	
	\noindent We aim to bound $\mathbb{E} \left[ \| \theta_{c,i} - \theta \|^2 \right]$, which is given by
	\begin{multline} \label{eq:drift-non-convex-1}
		\mathbb{E} \left[ \| \theta_{c,i} - \theta \|^2 \right] \\ \leq 4 \gamma^2 \mathbb{E} \left[ \left\| \sum_{k=1}^{c} \sum_{f=0}^{p_{c,i}(k)} \left( \rvq_{\psi_k,f} - \nabla \mathcal{L}_{\psi_k} (\theta_{k,f}) \right) \right\|^2 \right] \\
		 + 4 \gamma^2 \mathbb{E} \left[ \left\| \sum_{k=1}^{c} \sum_{f=0}^{p_{c,i}(k)} \left( \nabla \mathcal{L}_{\psi_k} (\theta_{k,f}) - \nabla \mathcal{L}_{\psi_k} (\theta) \right) \right\|^2 \right] \\
		 + 4 \gamma^2 \mathbb{E} \left[ \left\| \sum_{k=1}^{c} \sum_{f=0}^{p_{c,i}(k)} \left( \nabla \mathcal{L}_{\psi_k} (\theta) - \nabla \mathcal{L} (\theta) \right) \right\|^2 \right] \\
		 + 4 \gamma^2 \mathbb{E} \left[ \left\| \sum_{k=1}^{c} \sum_{f=0}^{p_{c,i}(k)} \nabla \mathcal{L} (\theta) \right\|^2 \right].
	\end{multline}
	Next, we bound each term in Eq.~\eqref{eq:drift-non-convex-1}, for the first term we have:
	\begin{multline}
	4 \gamma^2 \mathbb{E} \left[ \left\| \sum_{k=1}^{c} \sum_{f=0}^{p_{c,i}(k)} \left( \rvq_{\psi_k,f} - \nabla \mathcal{L}_{\psi_k} (\theta_{k,f}) \right) \right\|^2 \right] \\ \leq 4 \gamma^2 \sum_{k=1}^c \sum_{f=0}^{p_{c,i}(k)} \mathbb{E} \left[ \| \rvq_{\psi_k,f} - \nabla \mathcal{L}_{\psi_k} (\theta_{k,f}) \|^2 \right] \leq 4 \gamma^2 \gY_{c,i} \nu^2,
	\end{multline}
	\noindent For the second term in Eq. \eqref{eq:drift-non-convex-1} we have:
	\begin{multline}
		4 \gamma^2 \mathbb{E} \left[ \left\| \sum_{k=1}^{c} \sum_{f=0}^{p_{c,i}(k)} \left( \nabla \mathcal{L}_{\psi_k} (\theta_{k,f}) - \nabla \mathcal{L}_{\psi_k} (\theta) \right) \right\|^2 \right]\\ \leq 4 \gamma^2 \gY_{c,i} \sum_{k=1}^c \sum_{f=0}^{p_{c,i}(k)} \mathbb{E} \left[ \| \nabla \mathcal{L}_{\psi_k} (\theta_{k,f}) - \nabla \mathcal{L}_{\psi_k} (\theta) \|^2 \right] \\
		\leq 4 L^2 \gamma^2 \gY_{c,i} \sum_{k=1}^c \sum_{f=0}^{p_{c,i}(k)} \mathbb{E} \left[ \| \theta_{k,f} - \theta \|^2 \right],
	\end{multline}
	\noindent For the third term in Eq. \eqref{eq:drift-non-convex-1} we can write:
	\begin{multline} \label{eq:drift-non-convex-2}
		4 \gamma^2 \mathbb{E} \left[ \left\| \sum_{k=1}^{c} \sum_{f=0}^{p_{c,i}(k)} \left( \nabla \mathcal{L}_{\psi_k} (\theta) - \nabla \mathcal{L} (\theta) \right) \right\|^2 \right]  \\ \leq 4 \gamma^2 \gY_{c,i} \mathbb{E} \left[ \| \nabla \mathcal{L}_{\psi_k} (\theta) - \nabla \mathcal{L} (\theta) \|^2 \right],
	\end{multline}
	and finally for the fourth term in Eq. \eqref{eq:drift-non-convex-1} we have:
	\begin{equation}
		4 \gamma^2 \mathbb{E} \left[ \left\| \sum_{k=1}^{c} \sum_{f=0}^{p_{c,i}(k)} \nabla \mathcal{L} (\theta) \right\|^2 \right] \leq 4 \gamma^2 \gY_{c,i}^2 \| \nabla \mathcal{L} (\theta) \|^2.
	\end{equation}
	\noindent Suppose we denote $\gY_{c,i} = (c-1)I + i$, then by replacing into the Eq.~\eqref{eq:drift-non-convex-1}, we have:
	\begin{align*}
		E_j &\leq 4 \gamma^2 \nu^2 \sum_{c=1}^J \sum_{i=0}^{I-1} \gY_{c,i} \\ &+ 4 L^2 \gamma^2 \sum_{c=1}^J \sum_{i=0}^{I-1} \gY_{c,i} \sum_{k=1}^c \sum_{f=0}^{p_{c,i}(k)} \mathbb{E} \left[ \| \theta_{k,f} - \theta \|^2 \right] \\
		& + 4 \gamma^2 \sum_{c=1}^J \sum_{i=0}^{I-1} \gY_{c,i} + 4 \gamma^2 \sum_{c=1}^J \sum_{i=0}^{I-1} \gY_{c,i}^2 \| \nabla \mathcal{L} (\theta) \|^2.
	\end{align*}
	
	\noindent Applying the bounds $\sum_{c=1}^J \sum_{i=0}^{I-1} \gY_{c,i} \leq \frac{1}{2} J^2 I^2$ and $\sum_{c=1}^J \sum_{i=0}^{I-1} \gY_{c,i}^2 \leq \frac{1}{3} J^3 I^3$, we have:
	\begin{align*}
		E_j &\leq 2 J^2 I^2 \gamma^2 \sigma^2 + 2 L^2 J^2 I^2 \gamma^2 E_j + 2 J^2 I^3 \gamma^2 \nu^2 \\
		&\quad + 2 \left( \frac{\beta^2}{J} + 1 \right) J^3 I^3 \gamma^2 \| \nabla \mathcal{L} (\theta) \|^2.
	\end{align*}
	
	\noindent Manipulating and forcing the relation $\gamma \leq \frac{1}{6 L I J (1 + \beta^2 / J)}$, we reach the result:
	\begin{multline}
		E_j \leq \frac{9}{4} J^2 I^2 \gamma^2 \sigma^2 + \frac{9}{4} J^2 I^3 \gamma^2 \nu^2 \\ + \frac{9}{4} \left( \frac{\beta^2}{J} + 1 \right) J^3 I^3 \gamma^2 \| \nabla \mathcal{L} (\theta) \|^2.
	\end{multline}
	
	\noindent The claim follows after recovering the superscripts and taking unconditional expectations.
\end{proof}

\subsection{Tuning the learning rate}

In the next Lemma we drive a bound on the delay aware learning rate for the \afl~algorithm.

\begin{lemma}\label{learning_rate_tune}
	We consider two non-negative sequences \(\{d_t\}_{t \geq 0}\), \(\{g_t\}_{t \geq 0}\), which satisfy the relation
	\begin{align}
		d_{t+1} \leq d_t - b\zeta_t g_t + a_1 \zeta_t^2 + a_2 \zeta_t^3, \label{eq1:lem:dynamic_learning_rate_updated}
	\end{align}
	for all \(t \geq 0\), where \(b > 0\), \(a_1, a_2 \geq 0\), and \(\zeta_t\) is the learning rate defined as 
	\[
		\zeta_t = \frac{\zeta_0}{\sqrt{t+1} \cdot (1 + \alpha \cdot \tau_t)}.
	\]
	Then, using \(\pi_t = 1\) and \(\Pi_T \coloneqq \sum_{t=0}^T \pi_t = T+1\), we have:
\begin{multline*}
\Psi_T \coloneqq \frac{1}{\Pi_T} \sum_{t=0}^T g_t \pi_t \leq \frac{d_0}{b\zeta_0(T+1)} \left(\sqrt{T+1} + \alpha \sum_{t=0}^T \tau_t\right) \\ + \frac{a_1\zeta_0}{b} \cdot \frac{\log(T+1)}{\sqrt{T+1}} + \frac{a_2\zeta_0^2}{b} \cdot \frac{\log(T+1)}{T}.  
\end{multline*}
\end{lemma}

\begin{proof}
Rearranging Eq.~\eqref{eq1:lem:dynamic_learning_rate_updated} and multiplying both sides by \(\pi_t = 1\), we get:
\[
	b g_t \leq \frac{d_t}{\zeta_t} - \frac{d_{t+1}}{\zeta_t} + a_1 \zeta_t + a_2 \zeta_t^2.
\]
Summing over \(t=0\) to \(T\) results in:
\[
	b \sum_{t=0}^T g_t \leq \sum_{t=0}^T \frac{d_t - d_{t+1}}{\zeta_t} + a_1 \sum_{t=0}^T \zeta_t + a_2 \sum_{t=0}^T \zeta_t^2.
\]
The first term forms a telescoping sum:
\[
	\sum_{t=0}^T \frac{d_t - d_{t+1}}{\zeta_t} = \frac{d_0}{\zeta_0} \left(\sqrt{T+1} + \alpha \sum_{t=0}^T \tau_t \right).
\]
For the second term, substituting \(\zeta_t = \frac{\zeta_0}{\sqrt{t+1} \cdot (1 + \alpha \cdot \tau_t)}\), we use bounds:
\[
	\sum_{t=0}^T \zeta_t \leq \zeta_0 \sum_{t=0}^T \frac{1}{\sqrt{t+1}} \leq \zeta_0 (2\sqrt{T+1} - 1).
\]
Similarly, for \(\zeta_t^2\):
\[
	\sum_{t=0}^T \zeta_t^2 \leq \zeta_0^2 \sum_{t=0}^T \frac{1}{t+1} \leq \zeta_0^2 \log(T+1).
\]
Dividing by \(\Pi_T = T+1\), we have:
\begin{multline*}
\Psi_T \leq \frac{d_0}{b\zeta_0(T+1)} \left(\sqrt{T+1} + \alpha \sum_{t=0}^T \tau_t \right) \\ + \frac{a_1\zeta_0}{b} \cdot \frac{\log(T+1)}{\sqrt{T+1}} + \frac{a_2\zeta_0^2}{b} \cdot \frac{\log(T+1)}{T}.
\end{multline*}
The first term scales as \( O\left(\frac{1}{\sqrt{T}}\right) \) when \( \tau_t = 0 \) (homogeneous case). The second and third terms decay at \( O\left(\frac{\log(T)}{\sqrt{T}}\right) \) and \( O\left(\frac{\log(T)}{T}\right) \), respectively, showing that higher-order terms become negligible for large \(T\). Thus, the result reveals that the average gradient norm \( \Psi_T \) decreases asymptotically as \( O(1/\sqrt{T}) \), assuming a decaying learning rate schedule.

\end{proof}


\begin{theorem}\label{conv_theorem}
Assume that all the local objective functions are \(L\)-smooth (Assumption~\ref{l_smooth}) and non-convex. For the \afl~algorithm (Algorithm~\ref{afl_non_convex_alg}), there exists a constant effective learning rate \(\tilde{\gamma} = \gamma CI\) and weights \(\{\pi_j\}_{j \geq 0}\) such that the weighted average of the global parameters  
\[
\bar{\theta}^{(\mathcal{J})} = \frac{\sum_{j=0}^{\mathcal{J}} \pi_j \theta^{(j)}}{\sum_{j=0}^\mathcal{J} \pi_j},
\]  

Under Assumptions~\ref{bounded_variance} (bounded variance) and \ref{asm:heterogeneity:optima} (bounded heterogeneity), if \(\tilde{\gamma} \leq \frac{1}{6L(1 + \beta^2 / C)}\) and \(\pi_j = 1\), then satisfies the following bound:

\begin{multline*}
\E\left[\|\nabla \mathcal{L}(\theta^{(j)})\|^2\right]  
\leq \frac{6 (\mathcal{L}(\theta^{(0)}) - \mathcal{L}^\ast) }{\tilde{\eta} \mathcal{J}} \\ + \frac{12L\tilde{\gamma}\sigma^2}{CI} + \frac{45L^2\tilde{\gamma}^2\sigma^2}{4CI} + \frac{45L^2\tilde{\gamma}^2\nu^2}{4C}.
\end{multline*}

\end{theorem}

\begin{proof}
From Lemmas~\ref{lem:non-convex:recursion} and~\ref{lem:non-convex:drift}, and assuming \(\gamma \leq \frac{1}{6LIJ(1+\beta^2/J)}\), the recursion simplifies to (see the Appendix \ref{appendix} for details):
\begin{align}\label{equation_appendix}
    \E \left[\mathcal{L}(\theta^{(j+1)}) - \mathcal{L}(\theta^{(j)})\right] 
    &\leq -\frac{1}{6}IJ\gamma \E\left[\|\nabla \mathcal{L}(\theta^{(j)})\|^2\right] 
    \\&+ 2L\gamma^2 IJ\sigma^2 
    + 2L\gamma^2 J^2 I^2 \frac{C-J}{J(C-1)} \nu^2 \\
    & + \frac{15}{8}L^2\gamma^3 J^2 I^2 \sigma^2 
    + \frac{15}{8}L^2\gamma^3 J^2 I^3 \nu^2.
\end{align}

\noindent Introducing the substitution \(\tilde{\gamma} \coloneqq IJ\gamma\), we rewrite the recursion after subtracting \(\mathcal{L}^\ast\) from both sides:
\begin{align*}
    \E \left[\mathcal{L}(\theta^{(j+1)}) - \mathcal{L}^\ast\right] 
    &\leq \E\left[\mathcal{L}(\theta^{(j)}) - \mathcal{L}^\ast\right] 
    \\ &- \frac{\tilde{\gamma}}{6} \E\left[\|\nabla \mathcal{L}(\theta^{(j)})\|^2\right] \\
    & + \frac{2L\tilde{\gamma}^2 \sigma^2}{IJ} 
    + 2L\tilde{\gamma}^2 \frac{C-J}{J(C-1)} \nu^2 
    \\& + \frac{15}{8}\frac{L^2\tilde{\gamma}^3 \sigma^2}{IJ} 
     + \frac{15}{8}\frac{L^2\tilde{\gamma}^3 \nu^2}{J}.
\end{align*}

Now, applying Lemma~\ref{learning_rate_tune}, we define the following parameters: \(t = j\), \(T = \mathcal{J}\), \(\zeta = \tilde{\gamma}\), \(d_t = \E\left[\mathcal{L}(\theta^{(j)}) - \mathcal{L}^\ast\right]\), \(b = \frac{1}{10}\), \(g_t = \E\left[\|\nabla \mathcal{L}(\theta^{(j)})\|^2\right]\), \(\pi_t = 1\), \(a_1 = \frac{2L\sigma^2}{IJ} + 2L\frac{C-J}{J(C-1)} \nu^2\), and \(a_2 = \frac{15}{8}\frac{L^2 \sigma^2}{IJ} + \frac{15}{8}\frac{L^2 \nu^2}{J}\). Using these substitutions, we derive:
\begin{align}
    \E\left[\|\nabla \mathcal{L}(\theta^{(j)})\|^2\right] 
    &\leq \frac{6\left(\mathcal{L}(\theta^{0}) - \mathcal{L}^\ast\right)}{\tilde{\gamma} \mathcal{J}} 
    + \frac{12L\tilde{\gamma} \sigma^2}{IJ}\nonumber \\ &
    + \frac{12L\tilde{\gamma} \nu^2 (C-J)}{J(C-1)} \nonumber \\
    &\quad + \frac{45L^2 \tilde{\gamma}^2 \sigma^2}{4IJ} 
    + \frac{45L^2 \tilde{\gamma}^2 \nu^2}{4J}. \label{eq:thm:non-convex:bound}
\end{align}

When \(J = C\), the result in the theorem statement follows. Specifically:
1. The first term, \(O(1/\mathcal{J})\), dominates the bound as the number of iterations increases.
2. The second term, \(O(\gamma)\), reflects variance (\(\sigma^2\)) and diminishes as \(\gamma\) decreases.
3. The remaining terms, \(O(\gamma^2)\), arise from both variance and heterogeneity (\(\nu^2\)), and scale quadratically with \(\gamma\).

Choosing a learning rate of \(\gamma \sim O(1/\sqrt{\mathcal{J}})\) ensures the expected gradient norm satisfies:
\[
\E\left[\|\nabla \mathcal{L}(\theta^{(j)})\|^2\right] \sim O\left(\frac{1}{\sqrt{\mathcal{J}}}\right),
\]
which is consistent with the established convergence rate for non-convex optimization.
\end{proof}

\section{Simulations}\label{simulation}
In this section we consider \afl~algorithm to tackle challenges related to heterogeneous client environments and the optimization of non-convex objective functions. We evaluate the performance of our \afl~algorithm using well-known benchmark datasets, namely MNIST and CIFAR, with non-convex optimization objectives. Additionally, we compare the results of the \afl~algorithm with those of a synchronous federated learning (FL) approach.

\subsection{Hardware Setup}
Our deep learning model training was conducted on high-performance hardware to meet the computational demands of wind power simulations. Specifically, we utilized an \texttt{NVIDIA A100 GPU} with 32 \texttt{GB} of Video Random Access Memory (\texttt{VRAM})\footnote{\texttt{VRAM} refers to the dedicated memory on a \texttt{GPU}, optimized for high-speed graphics processing, parallel computations, and deep learning tasks. Unlike standard \texttt{RAM}, which is used by the \texttt{CPU}, \texttt{VRAM} efficiently handles large-scale matrix operations, simulations, and neural network training.}. Additionally, each \texttt{CPU} core was allocated 256 \texttt{GB} of memory to facilitate large-scale data processing.

The training process was managed using the Simple Linux Utility for Resource Management (\texttt{SLURM}) job scheduler\footnote{\texttt{SLURM} is a widely used open-source workload manager for high-performance computing (\texttt{HPC}) environments. Developed by the Lawrence Livermore National Laboratory, it efficiently allocates resources such as \texttt{CPU}s, \texttt{GPU}s, and memory across computing nodes, enabling optimal job scheduling and execution.}. Job-specific constraints were applied to ensure compatibility with the \texttt{A100 GPU}'s extensive memory capacity, optimizing parallel processing and maintaining model training stability.

\subsection{Code and data availability statement.}
The \afl~algorithm is implemented using the \texttt{PyTorch} framework, taking advantage of its advanced tools for constructing and training deep learning models. It is openly available to enable reproducibility and facilitate further research at this \texttt{Github} repository \footnote{\url{https://github.com/Ali-Forootani/AFL_non_convex_non_iid}}. The code for data extraction, interpolation, statistical analysis, visualization, DNN~training and evaluation is provided in the associated repository. It includes detailed comments and instructions for reproducing the results.
Additionally, a version will be archived on \texttt{Zenodo} \footnote{\url{https://zenodo.org/records/14962410}} for reference.

\subsection{Classification task on MNIST dataset}
The dataset used in this experiment is the MNIST dataset, which contains $60,000$ training images and $10,000$ test images of handwritten digits. To simulate realistic federated learning conditions, the data was partitioned among clients in a non-IID manner using a Dirichlet distribution with a concentration parameter \(\alpha = 0.5\). A lower value of \(\alpha\) creates more uneven distributions, meaning some clients may have a disproportionate number of samples for specific classes. Each image is normalized and transformed into a PyTorch tensor for model input.

\paragraph{Training setup} The model used for both clients and the server is a Convolutional Neural Network (CNN). The network architecture consists of three convolutional layers, each followed by a ReLU activation function. The convolutional layers use $32$ filters with a kernel size of \(3 \times 3\), a stride of $1$, and padding of $1$, ensuring the spatial dimensions remain consistent across layers. The weights of the convolutional layers are initialized using Xavier uniform initialization to improve convergence. Following the convolutional layers is an adaptive average pooling layer, which reduces the spatial dimensions to \(1 \times 1\). A fully connected layer maps the resulting features to $10$ output classes, corresponding to the digits $0$ through $9$. The final output layer applies a log-softmax function, making the model suitable for multi-class classification.

The federated learning procedure follows the standard server-client communication protocol. At the start of each communication round, a subset of clients is selected randomly to participate. Each selected client trains its local model on its dataset for a specified number of epochs before sending the updated model weights back to the server. The server aggregates the client models using a weighted average, where the weight corresponds to the size of the client's local dataset. This aggregated model is then broadcast back to the clients for the next round of training.

The learning rate at each client is adjusted dynamically based on the delay using the formula \(\gamma_t = \frac{\gamma_0}{\sqrt{t + 1} \cdot (1 + \alpha \cdot d_t)}\), where \(\gamma_0\) is the initial learning rate, \(t\) is the epoch number, \(\alpha\) is the delay scaling factor, and \(d_t\) is the delay time. This ensures that clients with longer delays make smaller updates, preventing them from destabilizing the global model. An early stopping mechanism is implemented to terminate training if the client's loss does not improve for $10$ consecutive epochs, reducing unnecessary computation.

\texttt{PyTorch} is used for model definition and training, while the \texttt{asyncio} and \texttt{nest\_asyncio} packages enabled asynchronous client updates. The experiment is configured with $C=10$ clients, each training for $I=10$ local epochs per communication round, with a batch size of $64$. The total number of communication rounds is set to $\mathcal{J}=1000$, and in each round. The initial learning rate is set to \(\gamma_0 = 10^{-3}\), and the delay is computed dynamically in the code.

The loss function used for training is negative log-likelihood loss (\texttt{NLLLoss}), appropriate for multi-class classification tasks. The Adam optimizer \cite{Adam} is chosen for its adaptive learning rate capabilities, which complement the delay-aware adjustments.

\paragraph{Results}

\begin{figure}
    \centering
    \includegraphics[width=0.99\columnwidth]{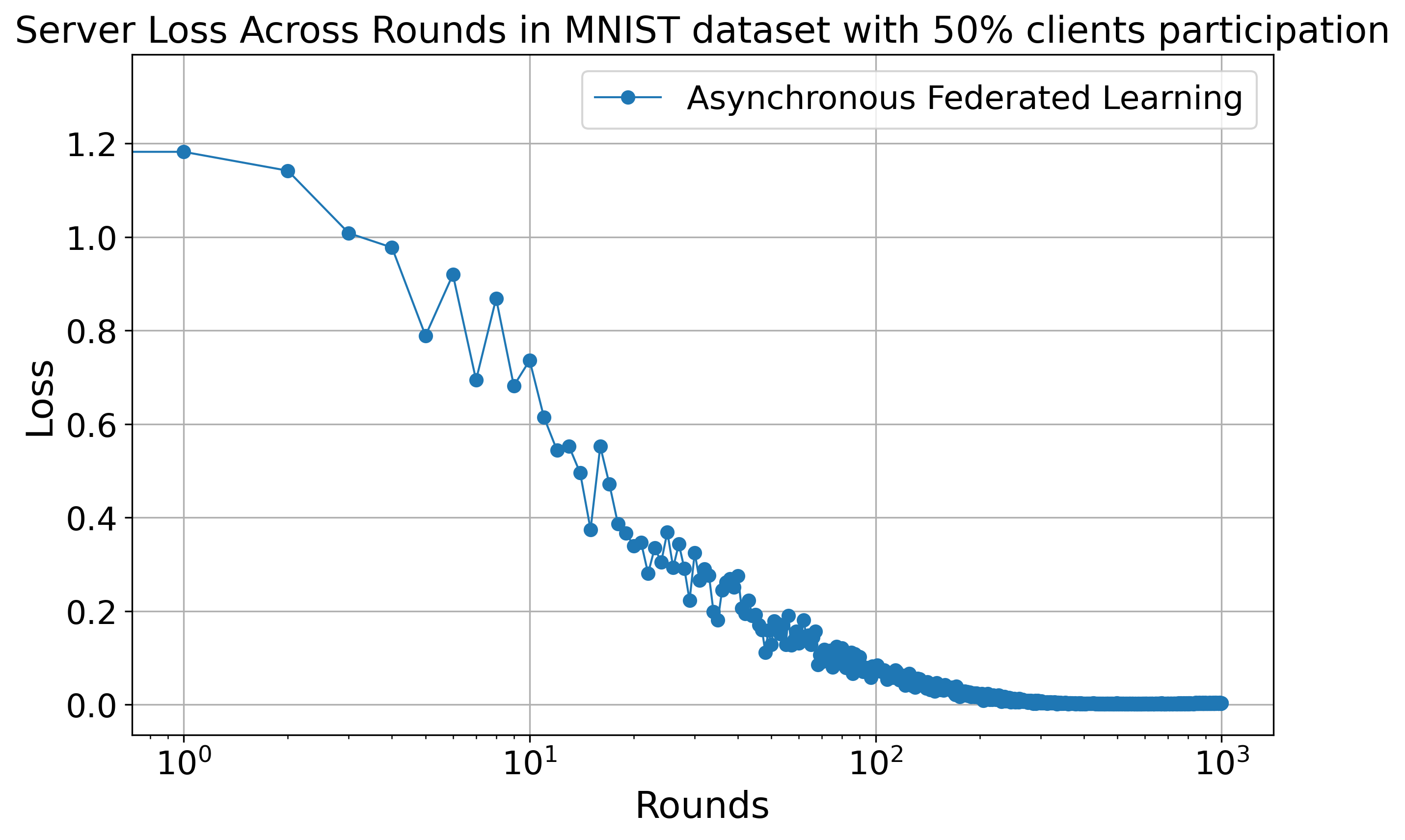}
    \caption{Server loss across rounds in \afl~algorithm for MNIST dataset with participation of $5$ clients in the training loop.}
    \label{afl_mnist_1}
\end{figure}

In the first experiment $J=5$ clients are randomly selected to participate in the training. The global server model demonstrated consistent improvement across the communication rounds as shown in figure \ref{afl_mnist_1}. The server loss decreased steadily, with diminishing returns toward the later rounds, indicating convergence. This trend suggests that the server is able to generalize well despite the heterogeneous data distributions and asynchronous training times of the clients.

The non-IID data distribution can introduce additional challenges, as clients with highly imbalanced datasets required more rounds to contribute effectively to the global model. Despite this, the model aggregation strategy is able to handle the variability in client updates, resulting in a robust and generalized server model.

\begin{figure}
    \centering
    \includegraphics[width=0.85\columnwidth]{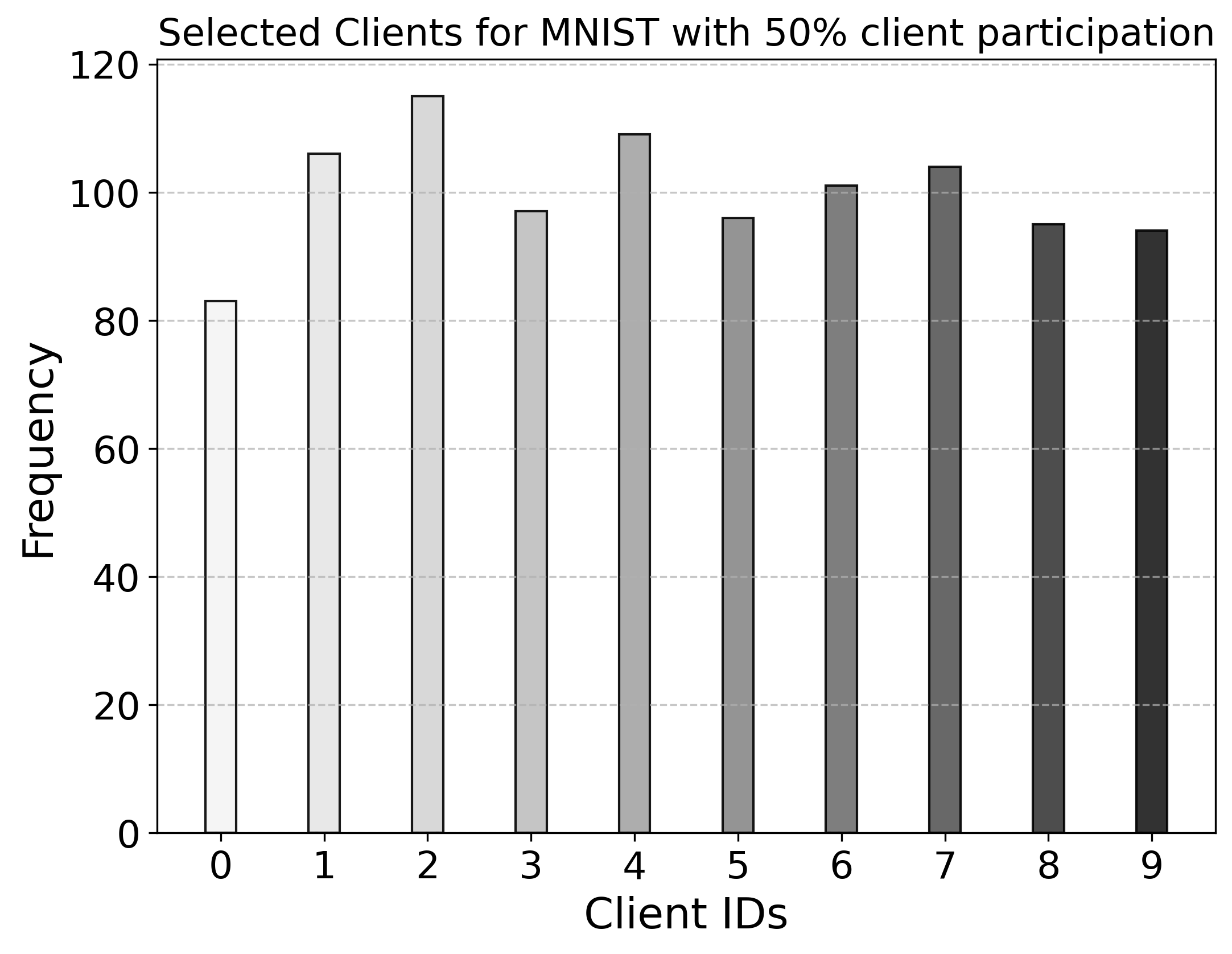}
    \caption{Frequency distribution of different clients during the training loop in \afl~algorithm with participation of $5$ clients in the training loop for the MNIST dataset.}
    \label{mnist_client_selection}
\end{figure}

Figure~\ref{mnist_client_selection} presents a histogram depicting the frequency of client selection in the federated learning process. The results indicate that some clients, such as client 2, have been selected more frequently than others. Despite imbalance participation of the clients in the training loop, the global convergence of the server is not affected significantly as shown in \ref{afl_mnist_1}.

\begin{figure}
    \centering
    \includegraphics[width=0.99\linewidth]{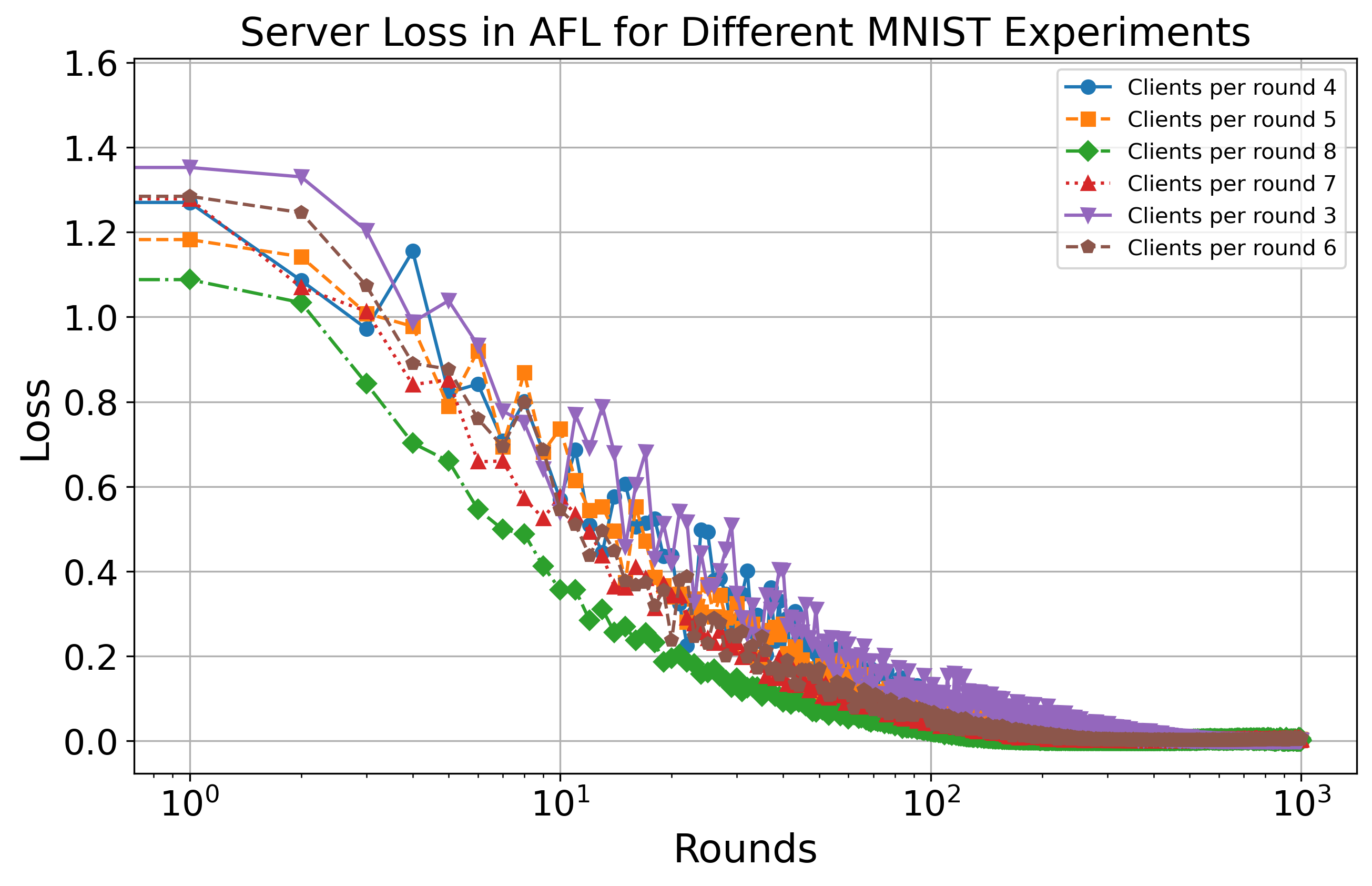}
    \caption{Server loss across rounds based on the various number of client participation in the \afl~algorithm for the classification task for the MNIST dataset.}
    \label{mnist_clients_participation}
\end{figure}

Figure \ref{mnist_clients_participation}, showing the server loss in the \afl~algorithm for different MNIST experiments, illustrates the impact of varying client participation on the training process. Different curves correspond to experiments where the number of clients participating per round ranges from 3 to 8.

From the plot \ref{mnist_clients_participation}, it is evident that increasing the number of clients per round leads to a faster reduction in server loss, indicating a more rapid convergence of the global model. As the number of clients per round increases, the frequency of model updates also rises, allowing the server to aggregate more information from diverse clients. This leads to faster improvement of the global model, especially in the early rounds of training.

However, after a certain point, the improvement in the server loss starts to diminish, as shown by the flattening of the curves. This indicates the onset of convergence, where further model updates yield smaller improvements. The results also suggest that, despite the asynchronous nature of the client updates, the server can still benefit from increased client participation. 

The variations in the curves highlight the balance between the speed of model convergence and the communication cost, as adding more clients per round can improve the server model more quickly but also increase the computational burden on the server to aggregate the larger number of updates. This trade-off is an important consideration when designing federated learning systems, especially in scenarios with high client variability, like the one presented with the MNIST dataset in your experiment.

\begin{figure}
    \centering
    \includegraphics[width=0.99\columnwidth]{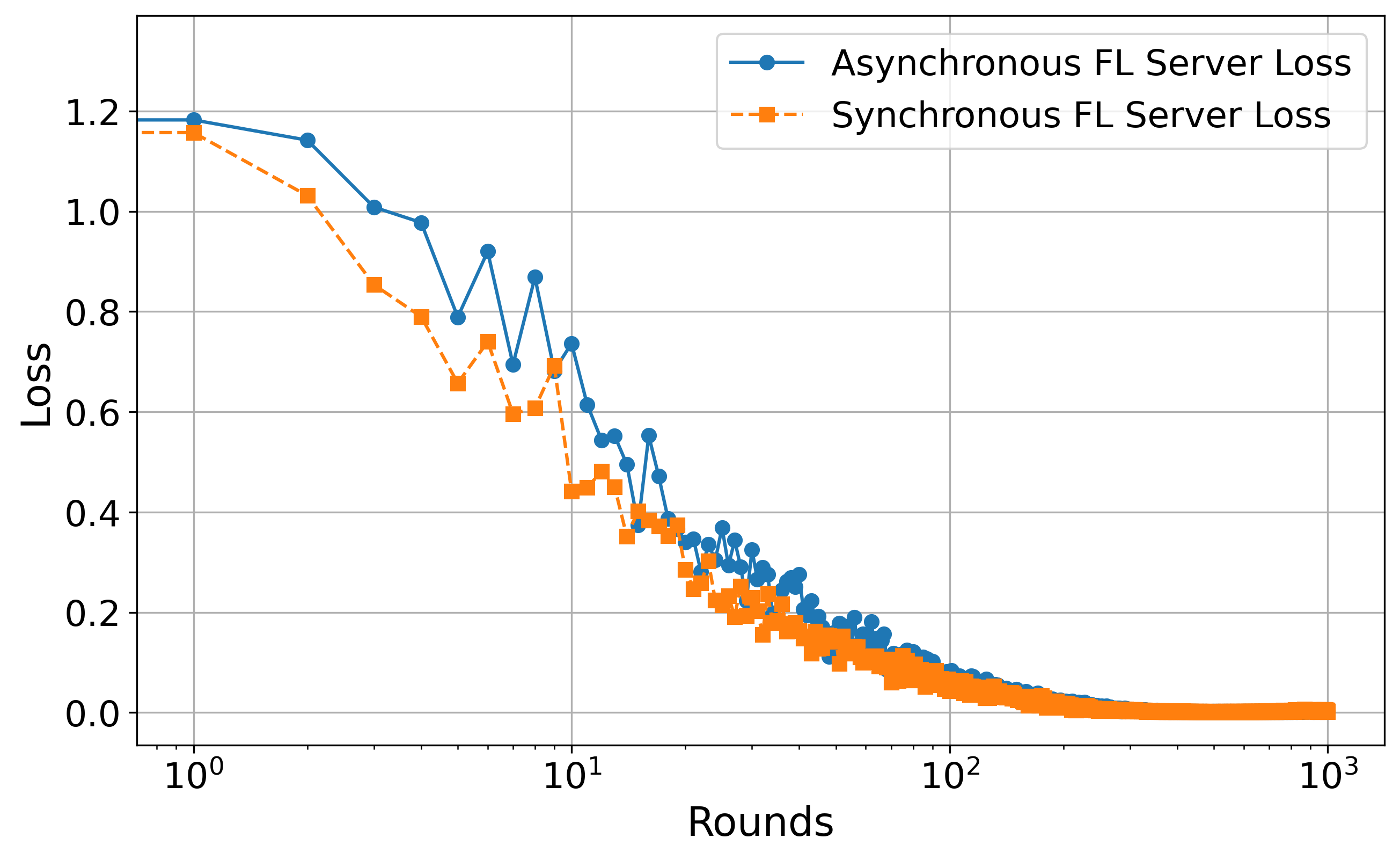}
    \caption{Comparison of the \afl~algorithm with Synchronous FL with participation of 5 clients in the training process.}
    \label{mnist_server_loss_comparison}
\end{figure}

Figure~\ref{mnist_server_loss_comparison} compares the training loss of asynchronous and synchronous federated learning (FL) on the MNIST dataset. Both approaches show a decreasing loss trend, confirming effective model training. However, synchronous FL exhibits a smoother loss reduction, while \afl~experiences higher variance, particularly in early training rounds. This fluctuation is likely due to staggered client updates and varying staleness in \afl~. Despite this variance, both methods eventually converge to similar loss levels, demonstrating that asynchronous FL remains a viable alternative to synchronous FL.


\subsection{Classification task on CIFAR dataset}
In the next example we employ the \afl~algorithm on the CIFAR-10 dataset using a ResNet architecture. CIFAR-10 contains 50{,}000 training and 10{,}000 test images in 10 classes, each image being $32\times 32$ pixels with three color channels (RGB). To reflect a realistic setup, the data is partitioned in a non-i.i.d.\ manner among 10 clients using a Dirichlet distribution with concentration parameter $\alpha=0.5$, so some clients may have data strongly skewed toward certain classes.
	
\paragraph{Training setup} Each client maintains a local ResNet model consisting of three layers of stacked blocks, where each block performs two $3\times 3$ convolutions with batch normalization and adds a shortcut path to mitigate vanishing gradients. After these layers, global average pooling reduces the spatial dimension, and a final fully connected layer outputs 10 logits \footnote{the term \emph{logits}, computed as the weighted sum of input features with an added bias, represent the raw decision scores before classification; in this dataset, they are thresholded at zero to assign binary labels.} followed by a log-softmax activation for classification. To mimic real-world conditions, client's delay are computed before training starts, and the code uses \texttt{asyncio} to handle these tasks concurrently.

A \textit{delay-aware learning rate} is adopted, scaling the base learning rate $\gamma_0 = 10^{-3}$ by
\( \frac{1}{\sqrt{t+1}\,\bigl(1 + \alpha \cdot d_t\bigr)},\) where $\alpha=0.01$ and $d_t$ is the client’s delay, so heavily delayed clients contribute smaller updates that reduce the risk of destabilizing the global model. In addition early stopping, with a patience of $10$ epochs, halts local training if the loss fails to improve, saving computation on less productive updates. Once local training finishes, the server aggregates the client's parameters via a weighted average proportional to each client’s dataset size, forming the updated global model.

\paragraph{Results}

At the first experiment in each communication round, a random subset of $J=5$ out of the $C=10$ clients is chosen to train locally for $I=10$ epochs, after which their models are sent back to a global server. Total number of training rounds is considered $\mathcal{J}=200$.

\begin{figure}
    \centering
    \includegraphics[width=0.99\columnwidth]{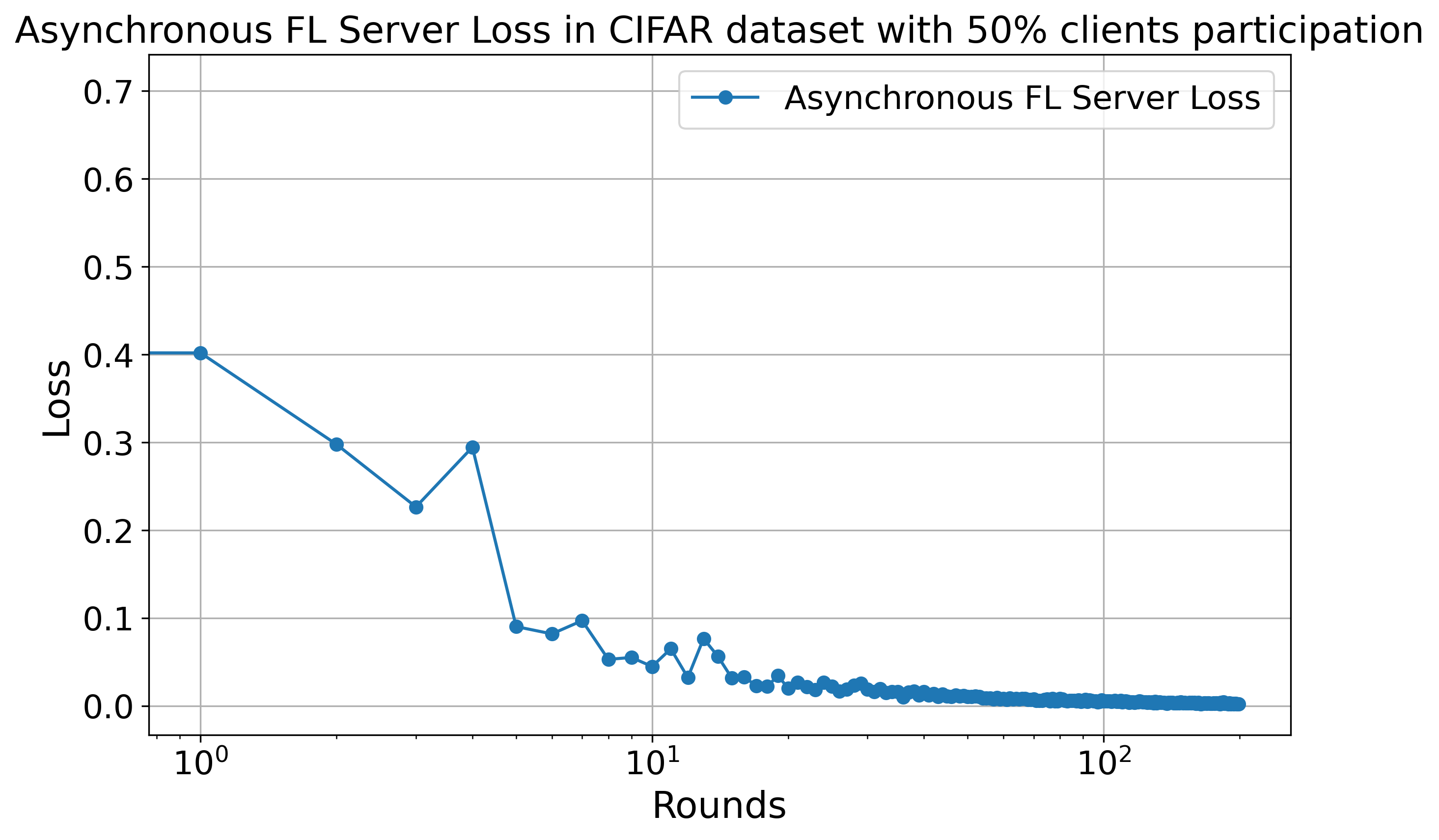}
    \caption{Server loss across rounds in \afl~algorithm for CIFAR dataset with participation of $5$ clients in the training loop.}
    \label{cifar_1}
\end{figure}

The simulation result for the first experiment is shown in the figure \ref{cifar_1}. Over $\mathcal{J}=200$ communication rounds, the server's average loss converges despite heterogeneous client distributions and client's delays, demonstrating the resilience of asynchronous federated learning with ResNet on CIFAR-10. In the early rounds, the server loss decreases significantly, suggesting that even with a small number of updates, the global model rapidly adapts to the distributed data.

The loss curve exhibits fluctuations at certain points, particularly in the first few rounds. This behavior is characteristic of \afl~, where updates arrive at different times, and clients contribute to the global model asynchronously.

\begin{figure}
    \centering
    \includegraphics[width=0.85\columnwidth]{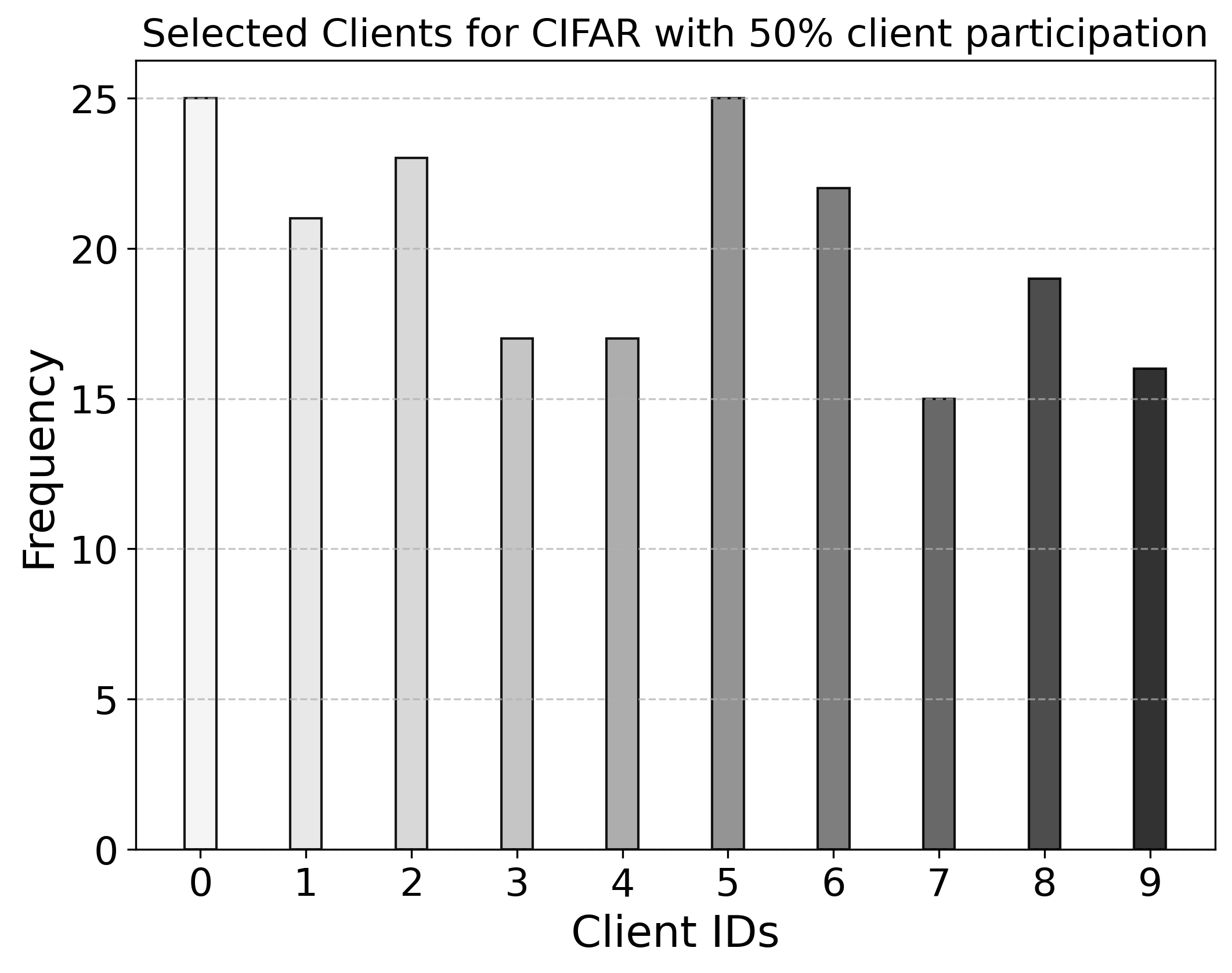}
    \caption{Frequency distribution of different clients during the training loop in \afl~algorithm with participation of $5$ clients in the training loop for the CIFAR dataset.}
    \label{cifar_client_selection}
\end{figure}

The histogram in \ref{cifar_client_selection} illustrates the frequency of client participation during the \afl~training process on the CIFAR-10 dataset, where 5 out of 10 clients are randomly selected per communication round. The variation in participation frequency among clients highlights the non-uniform selection process inherent in \afl~. Some clients, such as Client 0 and Client 5, are chosen more frequently than others, while some, like Client 7, have fewer training opportunities. This imbalance in client participation can introduce variability in model updates, yet the global model still converges effectively, as observed in the loss curve. The randomness in client selection ensures diversity in training data aggregation, helping the model generalize better despite the non-IID data distribution. These results demonstrate the robustness of \afl~in handling heterogeneous client availability while maintaining stable learning dynamics.

\begin{figure}
    \centering
    \includegraphics[width=0.99\columnwidth]{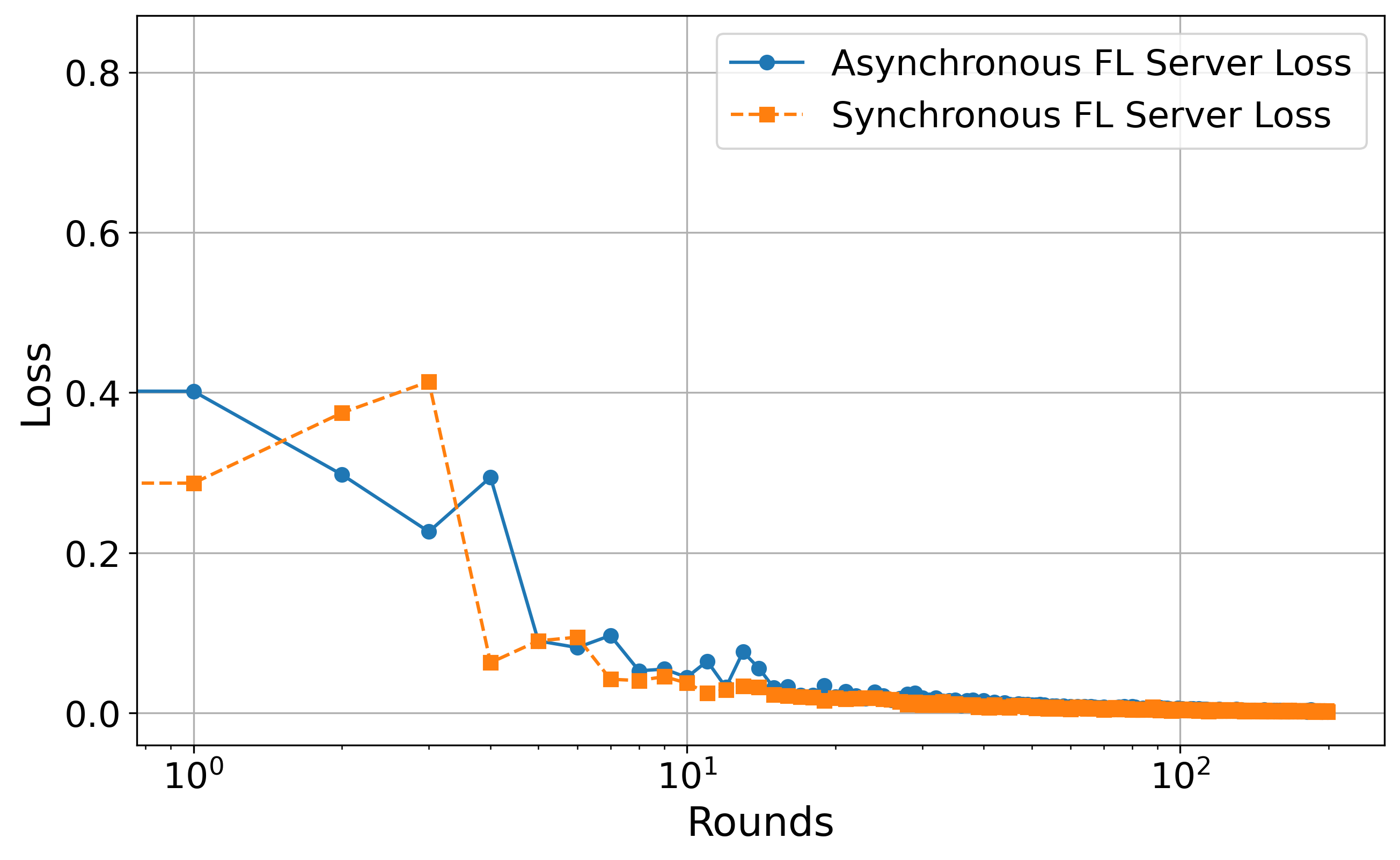}
    \caption{Comparison of the \afl~algorithm with Synchronous FL with participation 5 clients in the training loop for CIFAR dataset.}
    \label{cifar_server_loss_comparison}
\end{figure}

The comparison between \afl~and Synchronous FL on the CIFAR-10 dataset is shown in the figure \ref{cifar_server_loss_comparison} reveals key trade-offs in their training dynamics. While SFL ensures more stable updates by aggregating all selected client models before updating the global model, it introduces delays due to waiting for all clients. In contrast, \afl~enables faster updates by allowing clients to contribute asynchronously, resulting in more fluctuations in the early rounds but maintaining steady convergence over time.

\begin{figure}
    \centering
    \includegraphics[width=0.99\linewidth]{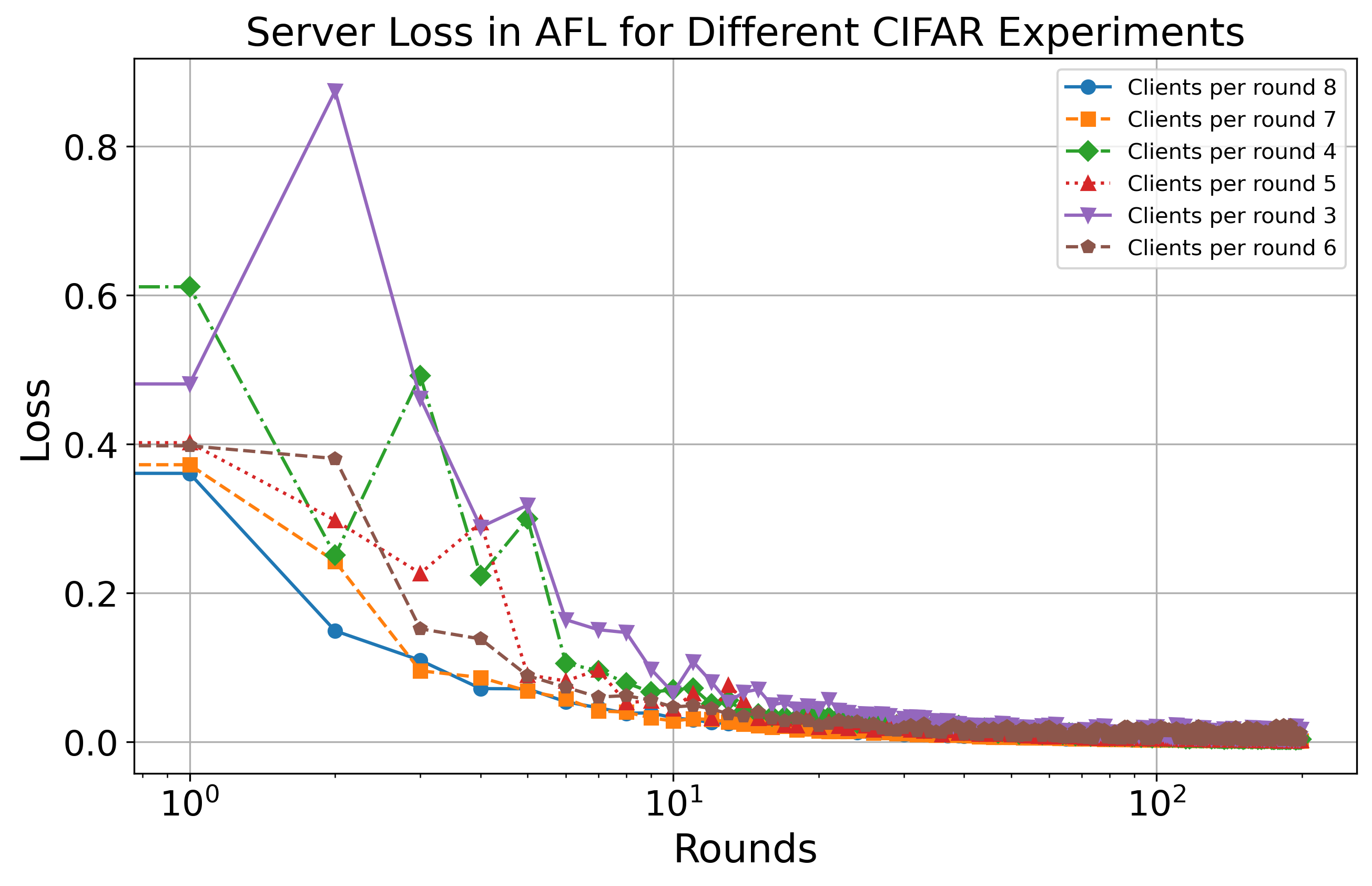}
    \caption{Server loss across rounds based on the various number of client participation in the \afl~algorithm for the classification task for the CIFAR dataset.}
    \label{cifar_clients_participation}
\end{figure}

Figure \ref{cifar_clients_participation} presents the impact of varying the number of participating clients per round on server loss in \afl~for the CIFAR-10 dataset. Each curve corresponds to a different number of participating clients per round, ranging from 3 to 8. From this figure, we observe that increasing the number of clients per round leads to a faster reduction in server loss, indicating improved convergence. When more clients participate, the global model receives more frequent and diverse updates, accelerating learning. However, the difference in convergence rates becomes less pronounced after a certain number of rounds, suggesting diminishing returns in increasing client participation. The results also highlight the robustness of \afl~in handling asynchronous updates, as all configurations eventually converge despite fluctuations in early rounds.

This experiment underscores the trade-off between efficiency and convergence speed in federated learning. While selecting more clients per round enhances learning stability, it also increases communication overhead. The results validate that even with a lower number of participating clients, \afl~remains effective, demonstrating its adaptability to real-world federated learning scenarios where client availability may be limited or unpredictable.

Despite initial variations, both \afl~and Synchronous FL achieve similar final loss values, demonstrating that \afl~can effectively learn from decentralized data without requiring strict synchronization. This makes \afl~a more communication-efficient alternative for federated learning settings where client availability varies.



\section{Conclusion} \label{conclusion}
In this paper, we extended asynchronous federated learning (\afl~) to the optimization of \textit{non-convex objective functions}, addressing key challenges such as client staleness, heterogeneity, and communication delays. Our approach introduced a \textit{staleness-aware aggregation mechanism} and a \textit{dynamic learning rate schedule} to adapt to the asynchronous nature of client updates, ensuring more stable convergence. The theoretical analysis provided mathematical bounds on the impact of stale gradients, demonstrating that our approach effectively mitigates their negative effects. Through extensive simulations on the MNIST and CIFAR-10 datasets, we validated the effectiveness of \afl~in handling \textit{non-IID data distributions} and \textit{heterogeneous client environments}. The results demonstrated that \afl~maintains robust convergence while reducing the impact of straggler clients, making it a scalable alternative to Synchronous FL. Specifically, our findings highlighted that increasing client participation per round accelerates convergence while preserving stability. The comparison between \afl~and Synchronous FL showed that despite early-round fluctuations, \afl~achieves similar final performance with lower synchronization overhead. This study contributes to the advancement of federated learning in practical, large-scale, and heterogeneous environments. Future work can focus on further refining staleness compensation mechanisms, incorporating adaptive sampling strategies, and exploring \afl~'s applicability to additional real-world tasks such as healthcare, finance, and smart grid systems.

\section{Appendix}\label{appendix}
This section provides more details related to the intermediate steps for derivation of Eq.\eqref{equation_appendix} in Theorem \ref{conv_theorem}. Starting from the inequality given in Lemma~\ref{lem:non-convex:recursion}:

\begin{multline}
\E\left[\mathcal{L}(\theta^{(j+1)}) - \mathcal{L}(\theta^{(j)})\right] \leq -\frac{1}{6}JI\gamma\E\left[\Norm{\nabla \mathcal{L}(\theta^{(j)})}^2\right] 
\\+ 2L\gamma^2JI\sigma^2 
+ 2L\gamma^2J^2I^2\frac{C-J}{J(C-1)}\nu^2 
+ \frac{5}{6}L^2\gamma E_j,  
\end{multline}

and substituting \(E_j\) from Lemma~\ref{lem:non-convex:drift}:

\begin{multline*}
E_j \leq \frac{9}{4}J^2I^2\gamma^2\sigma^2 
+ \frac{9}{4}J^2I^3\gamma^2\nu^2 
\\ + \frac{9}{4}\left(\frac{\beta^2}{J} + 1\right)J^3I^3\gamma^2\E\left[\Norm{\nabla \mathcal{L}(\theta^{(j)})}^2\right],
\end{multline*}

we substitute \(E_j\) into the recursion to get:

\begin{multline*}
\E\left[\mathcal{L}(\theta^{(j+1)}) - \mathcal{L}(\theta^{(j)})\right] 
\leq -\frac{1}{6}JI\gamma\E\left[\Norm{\nabla \mathcal{L}(\theta^{(j)})}^2\right] 
\\ + 2L\gamma^2JI\sigma^2 
\\+ 2L\gamma^2J^2I^2\frac{C-J}{J(C-1)}\nu^2 
+ \frac{15}{8}L^2\gamma^3J^2I^2\sigma^2 
+ \frac{15}{8}L^2\gamma^3J^2I^3\nu^2 
\\+ \frac{15}{8}L^2\gamma^3\left(\frac{\beta^2}{J} + 1\right)J^3I^3\E\left[\Norm{\nabla \mathcal{L}(\theta^{(j)})}^2\right].  
\end{multline*}

Focusing on the gradient-related terms:
\begin{multline*}
-\frac{1}{6}JI\gamma\E\left[\Norm{\nabla \mathcal{L}(\theta^{(r)})}^2\right] 
\\ + \frac{15}{8}L^2\gamma^3\left(\frac{\beta^2}{J} + 1\right)J^3I^3\E\left[\Norm{\nabla \mathcal{L}(\theta^{(r)})}^2\right].   
\end{multline*}

Substituting \(L = \frac{1}{6\gamma JI\left(1 + \frac{\beta^2}{J}\right)}\), we rewrite \(L^2\) as: \(
L^2 = \frac{1}{36\gamma^2J^2I^2\left(1 + \frac{\beta^2}{J}\right)^2}.\) Substituting \(L^2\) into the second term:

\begin{multline*}
\frac{15}{8}L^2\gamma^3\left(\frac{\beta^2}{J} + 1\right)J^3I^3 
\\ = \frac{15}{8} \cdot \frac{\gamma^3}{36\gamma^2J^2I^2\left(1 + \frac{\beta^2}{J}\right)^2} \cdot \left(\frac{\beta^2}{J} + 1\right) J^3I^3\\
= \frac{15}{8} \cdot \frac{\gamma JI}{36\left(1 + \frac{\beta^2}{J}\right)} \E\left[\Norm{\nabla \mathcal{L}(\theta^{(r)})}^2\right]
\end{multline*}



Combining with the first term:
\[
JI\gamma\E\left[\Norm{\nabla \mathcal{L}(\theta^{(r)})}^2\right] \left(-\frac{1}{6} + \frac{5}{96\left(1 + \frac{\beta^2}{J}\right)}\right).
\]

Simplify the coefficient:
\[
-\frac{1}{6} + \frac{5}{96\left(1 + \frac{\beta^2}{J}\right)} = \frac{-16\left(1 + \frac{\beta^2}{J}\right) + 5}{96\left(1 + \frac{\beta^2}{J}\right)} = \frac{-11 - \frac{16\beta^2}{J}}{96\left(1 + \frac{\beta^2}{J}\right)}.
\]

To provide a suitable upper bound for this coefficient, note that: \(
\frac{-11 - \frac{16\beta^2}{J}}{96\left(1 + \frac{\beta^2}{J}\right)} \leq -\frac{1}{6}. \) Thus, the final simplified recursion is:
\begin{multline*}
\E\left[\mathcal{L}(\theta^{(r+1)}) - \mathcal{L}(\theta^{(r)})\right] 
\leq -\frac{1}{6}JI\gamma\E\left[\Norm{\nabla \mathcal{L}(\theta^{(r)})}^2\right] \\+ \text{higher-order terms}.
\end{multline*}



\bibliographystyle{ieeetr}
\bibliography{my_refs_lcss,refs}


\end{document}